\documentclass[12pt]{article}

\usepackage{hyperref}
\usepackage{newtxtext,newtxmath}

\usepackage{graphicx}

\usepackage[letterpaper,margin=1in]{geometry}

\renewenvironment{abstract}
	{\quotation}
	{\endquotation}

\date{}

\makeatletter
\renewcommand{\fnum@figure}{\textbf{Figure \thefigure}}
\renewcommand{\fnum@table}{\textbf{Table \thetable}}
\makeatother

\usepackage{scicite}

\usepackage{url}

\providecommand{\bm}[1]{\boldsymbol{#1}}
\newcommand{\cmark}{\checkmark}
\newcommand{\xmark}{\texttimes}  
\newcommand{\npoints}{N_{\mathcal{P}}}

\def\scititle{
	Object-centric Task Representation and Transfer using Diffused
Orientation Fields}

\title{\bfseries \boldmath \scititle}

\author{
	Cem Bilaloglu$^{1,2\ast}$,
	Tobias Löw$^{1,2}$,
	Sylvain Calinon$^{1,2}$\and
	\small$^{1}$Idiap Research Institute, 1920 Martigny, Switzerland, \and
	\small$^{2}$EPFL, 1015 Lausanne, Switzerland\and 
	\small$^\ast$Corresponding author. Email: cem.bilaloglu@epfl.ch\and
}

\begin{document} 

\maketitle

\begin{abstract}
Curved objects pose a fundamental challenge for skill transfer in robotics: unlike planar surfaces, they do not admit a global reference frame. As a result, task-relevant directions such as “toward” or “along” the surface vary with position and geometry, making object-centric tasks difficult to transfer across shapes. To address this, we introduce an approach using Diffused Orientation Fields (DOF), a smooth representation of local reference frames, for transfer learning of tasks across curved objects. By expressing manipulation tasks in these smoothly varying local frames, we reduce the problem of transferring tasks across curved objects to establishing sparse keypoint correspondences. DOF is computed online from raw point cloud data using diffusion processes governed by partial differential equations, conditioned on keypoints. We evaluate DOF under geometric, topological, and localization perturbations, and demonstrate successful transfer of tasks requiring continuous physical interaction such as inspection, slicing, and peeling across varied objects. We provide our open-source codes at our website \url{https://github.com/idiap/diffused_fields_robotics}

\end{abstract}
\noindent


\section{Introduction}
    Humans routinely interact with curved objects as part of daily life, for example, slicing a banana, peeling a cucumber, or washing dishes. These tasks require continuous physical interaction with the object's surface, often involving making and breaking contact at different regions. In such tasks, motion on (or near) the surface is guided by the pose and geometry of the object. We refer to these interactions as object-centric tasks. Unlike collision avoidance, where simplified approximations are often sufficient, object-centric manipulation that involve physical interactions demand representations that accurately capture surface geometry and exploit its structure. A core challenge in these tasks is the immense variability in object shape. Even within a single category—such as cups or bananas—geometries can differ significantly in curvature, topology, and proportions. This variability makes it infeasible to learn and store manipulation strategies for every possible shape. 
    
    Although object-centric tasks are particularly challenging on complex surfaces, humans demonstrate a remarkable ability to transfer learned skills across objects with significantly different geometries. Once a skill is acquired, people can intuitively adapt their motion to new instances, even when the underlying shape changes substantially. We argue that this adaptability arises from using shape-invariant task descriptions paired with shape-aware representations that conform to the specific geometry of the object. This also directly impacts what information should be learned and transferred~\cite{jaquierTransferLearningRobotics2023}. For example, by learning trajectories using a fixed reference frame, there is the risk of overfitting both pose and object-specific details that can act as noise and hinder generalization. Such pose and shape-dependent details should indeed not be transferred to novel objects.

    A straightforward approach for pose-invariant skill representation is to define motions in the object’s local reference frame~\cite{urecheTaskParameterizationUsing2015}. More generally, encoding demonstrations from multiple reference frames enables extraction of task-relevant variability~\cite{calinonTutorialTaskparameterizedMovement2016}. Using multiple reference frames has been extended to a dictionary of coordinate systems including cylindrical and spherical frames~\cite{tiGeometricOptimalControl2023}, allowing selection of optimal geometric perspectives. Alternatively, coordinate-free descriptors~\cite{vochtenGeneralizingDemonstratedMotion2019, vochtenInvariantDescriptorsMotion2023,soCITRCoordinateInvariantTask2024} aim to capture invariants of motion without referring to coordinate systems.  However, these pose-invariant approaches do not consider the geometry of the object and typically overlook inter-category shape variation.

    In contrast, keypoint-based methods~\cite{manuelliKPAMKeyPointAffordances2019} describe manipulation tasks using geometric costs and constraints on selected landmarks. These offer robustness to intra-category variation and surface topology changes by filtering out irrelevant geometry. Extensions such as oriented keypoints combine local reference frames with keypoints to support action specification and closed-loop policies \cite{gaoKPAM20Feedback2021} and more recent methods leveraged keypoints for visual imitation learning~\cite{gaoKVILKeypointsBasedVisual2023, gaoBiKVILKeypointsbasedVisual2024}. Neural Descriptor Fields~\cite{simeonovNeuralDescriptorFields2022,simeonovSE3EquivariantRelationalRearrangement2022} go further by encoding local task descriptors into pose-invariant neural fields, while recent work leverages foundation models for learning task descriptors~\cite{wangD$^3$FieldsDynamic3D2024a}. Instead, policy learning approaches learn perception-to-action mapping from sensory inputs such as RGB~\cite{chiDiffusionPolicyVisuomotor2023} or point clouds~\cite{ze3DDiffusionPolicy2024}. Recent reactive-compliance variants explicitly integrate tactile/force feedback to stabilize contact~\cite{houAdaptiveCompliancePolicy2025, xueReactiveDiffusionPolicy}. However, these approaches are fundamentally data-driven—requiring demonstrations and offline training—and aim to learn end-to-end policies rather than a modular, geometry-aware representation that can be reused across objects, controllers and tasks.

    As an alternative to learning-based approaches, geometric methods use discrete differential geometry for task transfer across surfaces. Functional mapping methods~\cite{ovsjanikovFunctionalMapsFlexible2012,defariasGraspTransferDeformable2022,fariasGeometricallyAwareOneShotSkill2025} enable transferring functions over near-isometric surfaces given sparse correspondences; however, they are limited to open-loop position trajectories constrained to object surfaces. Other approaches use signed distances and surface parameterizations for ultrasound scanning~\cite{dyckImpedanceControlArbitrary2022}, approximate exponential maps for learning from demonstration on meshes~\cite{vedoveMeshDMPMotionPlanning2024}, and logarithmic maps for interactive grasping design~\cite{lakshmipathyContactEditArtist2023}. While effective within their scopes, these methods generally assume clean meshes, making them less suited to noisy, partial point clouds collected at runtime. Moreover, they do not provide a controller-agnostic representation that can be integrated as a module within other planning, learning, or control frameworks. Other efforts targeting continuous physical interaction on curved surfaces address specific problems (e.g., bimanual coordination~\cite{urecheConstraintsExtractionAsymmetrical2018} or force generation~\cite{amanhoudDynamicalSystemApproach2019}) but do not provide fine-grained adaptation to novel object instances.

    A key insight is that object-centric tasks like cleaning or slicing on a planar surface can often be reduced to simple, repetitive motions—such as up/down or back/forth—thanks to the presence of a consistent global frame. In contrast, curved or irregular objects lack such a global reference frame, making task specification and transfer significantly more challenging. Oriented keypoints~\cite{gaoKPAM20Feedback2021} or using multiple body-fixed reference frames~\cite{calinonTutorialTaskparameterizedMovement2016} can provide geometric structure for local interactions, however they are insufficient for continuous interactions with breaking and making contact near the surface. For that purpose, we construct a smoothly varying continuous representation of local frames throughout the robot's workspace conditioned on surface geometry and a few semantic keypoints. These local frames act as a geometric scaffold for expressing manipulation actions. For instance, a slicing or probing task might be described using a pose- and shape-invariant task description \emph{``slide along the object, go down and go up''}: an instruction that remains well-defined across object instances, categories, and poses. Therefore, our approach to represent and transfer object-centric tasks consists of (i) Diffused Orientation Fields (DOF) representing a smooth field of local reference frames conditioned on the object geometry and (ii) shape-invariant local actions defined in those frames. These local actions can be generated by various high-level controllers ranging from human in teleoperation to trajectory optimizer in planning and a policy programmed using primitives or learned using reinforcement learning. 

\begin{figure}[tbp]
	\centering
	\includegraphics[width=1.0\linewidth]{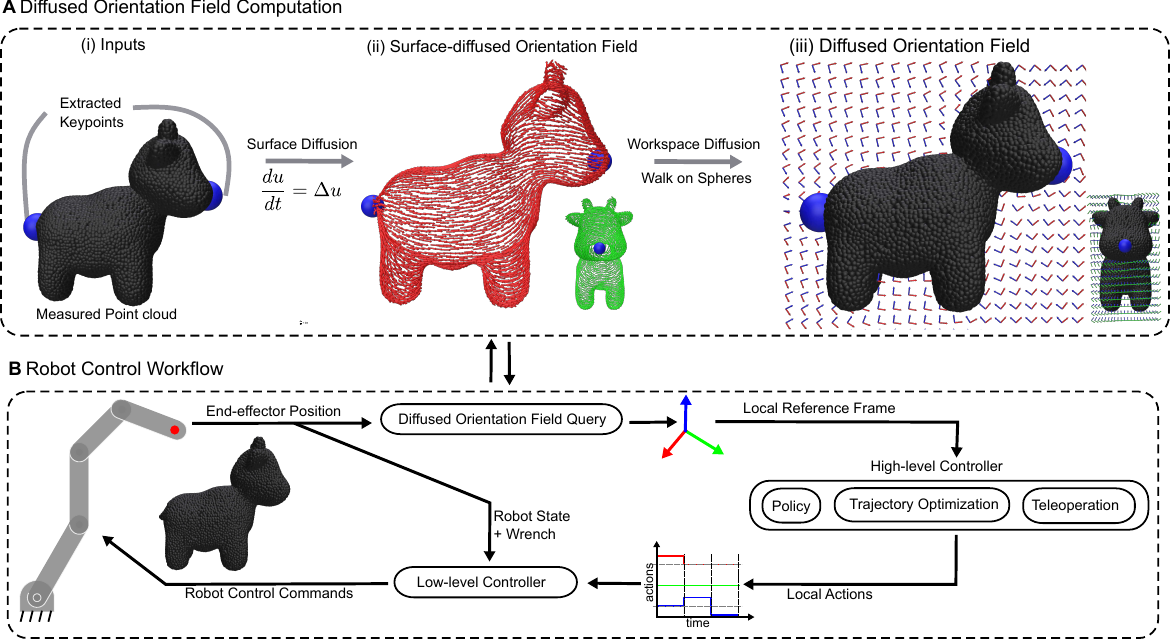}
	\caption{\textbf{Overview of the proposed representation with complete workflow}. We use Spot~\cite{craneRobustFairingConformal2013} as our canonical point cloud throughout the paper as it provides a sufficiently complex object with a well-defined symmetry axis for cross-section views. (\textbf{A}) Overview of the DOF computation.
    (i) Our method takes a point cloud collected at runtime and keypoints as its input. (ii) We compute a surface orientation field conditioned on the keypoints by solving the diffusion PDE on the point cloud. We visualize the orientation field by using local reference frames.  We show the x-axis in red, the y-axis in green and we omit the z-axis for clarity. (iii) We extend the surface orientation field to any point in the workspace using workspace diffusion. DOF represents smoothly varying local reference frames across the workspace by considering the object's surface geometry. We visualize it as a grid to show its smoothness, but in practice, we evaluate the field only at the query position. (\textbf{B}) Overview of our robot control workflow using local reference frames and actions. High-level controllers query the DOF at the robot position to obtain the local reference frame and produce local actions expressed in that frame for downstream tasks. These local actions are provided as references for the low-level tracking controller.}
	\label{fig:overview}
\end{figure}

    Our method reduces task-transfer to a novel object to computing its DOF from online vision and depth input while reusing the same shape-invariant local actions. This requires the representation to encode task-relevant geometric information across the workspace, support online computation and tolerate noisy and partial inputs. To construct DOF online, we draw inspiration from computer graphics. We use a partial differential equation (PDE) called the diffusion equation (also known as the heat equation), which provides a principled way to extend scalar and vector quantities over surfaces~\cite{craneGeodesicsHeatNew2013, sharpVectorHeatMethod2019}. The diffusion PDE is agnostic to surface discretization~\cite{sharpDiffusionNetDiscretizationAgnostic2022} and can be applied directly to point clouds~\cite{sharpLaplacianNonmanifoldTriangle2020}, making them well-suited to robotic problems where we do not always have access to watertight meshes of the objects. However prior work has focused on scalar and vector fields, and does not address the diffusion of orientations, which are critical for defining local reference frames. Moreover, many object-centric skills are not constrained to the object surface. Actions such as scooping or peeling often begin in free space and transition into contact. Classical PDE-based approaches require discretizing the entire workspace~\cite{fengHeatMethodGeneralized} which is not efficient. Although, methods like Kelvin-transformation~\cite{muchachoAdaptiveDistanceFunctions} provide efficient alternatives still they are impractical in robotics due to dynamic object positions. In contrast, Monte Carlo methods like Walk on Spheres (WoS)~\cite{sawhneyMonteCarloGeometry,muchachoWalkSpheresPDEbased2024} offer grid-free alternatives, but are currently limited to Euclidean domains and do not directly extend to curved surfaces. To overcome these limitations, we introduce a novel formulation for diffusing orientations that combines PDE and Monte Carlo-based methods summarized in Figure~\ref{fig:overview}.  Our contributions can be summarized as follows:
    \begin{itemize}
        \item expressing object-centric manipulation tasks as shape-invariant actions in local reference frames
        \item representing local reference frames as a smooth orientation field conditioned on the point cloud and keypoints collected online
        \item formulating the smooth orientation field using diffusion processes
        \item discretization-free online computation of the orientation field by combining surface and workspace diffusion processes
    \end{itemize}

    
    We demonstrate that our approach provides a shape-invariant representation for continuous, contact-rich tasks such as peeling, coverage, and slicing, enabling transfer to novel objects at runtime. The representation is computed online, robust to input noise, and modular to be integrated with different control, planning, or learning frameworks.

\section{Results}
\label{sec:experiments}

We provide an overview of our method and results in Movie~1.  Our experiments address a central question: can our method reduce the complexity of control, planning, and learning on curved objects and enable online transfer? To answer this, we organize the results into four sections:
\begin{enumerate}
    \item \textbf{Object-centric task representation and transfer}. We test DOF's ability to provide a shape-invariant representation that enables online transfer to novel objects. We consider three object-centric tasks: peeling, slicing and coverage. We implemented these tasks as local action primitives in local frames represented using DOF (Fig.~\ref{fig:skill_transfer}). Each task demonstrates a different tool to object relation: on the surface, above the surface, or penetrating the surface. We quantify variation in action statistics across diverse shapes and compare against baselines (Fig.~\ref{fig:transfer_comparison}).

    \item \textbf{Integration with different control paradigms}. We show the modularity of DOF as a geometric representation that can be integrated with different control, planning, and learning frameworks. We demonstrate DOF in reactive settings using closed-loop policies and teleoperation, in long-horizon settings using trajectory optimization, and with learned policies using reinforcement learning (Fig.~\ref{fig:controller_transfer}).
    
    \item \textbf{Robustness to sensing imperfections}. Since DOF is computed online from sensor data, we measure its robustness to geometric, keypoint and topological noise. We vary noise levels and occlusion and compare to the noise-free reference trajectory (Fig.~\ref{fig:robustness}). 

    \item \textbf{Multi-object scenes and compositionality}. We show DOF's use in complex scenes containing multiple geometric objects with different surface representations and compose a long-horizon scoop-lift-pour task using geometric primitives as constraints (Fig.~\ref{fig:primtives}).
\end{enumerate}

We provide a qualitative comparison of DOF against commonly used geometric representations in robotics (Table~\ref{tab:representation_comparison}) and present additional results for computational complexity and to alternative geometric baselines in the \emph{Supplementary Materials}.
\begin{figure}[]
\centering
\includegraphics[width=0.7\linewidth]{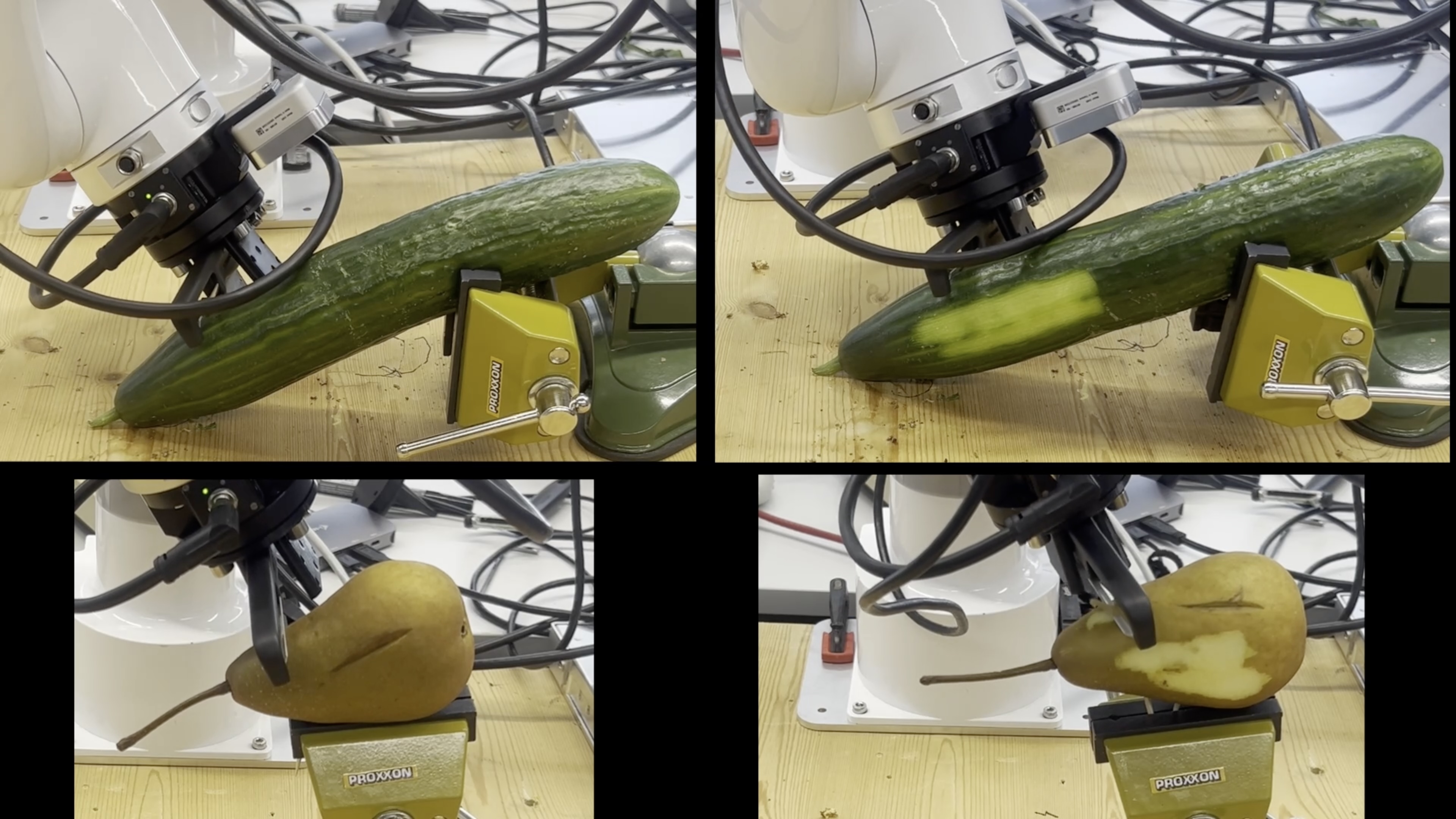}
\caption{\textbf{Moive 1: Summary of our method and results}. The accompanying video demonstrates our approach across tasks, objects, surface representations, and controllers in the real-world. We also show how DOFs are constructed from point clouds and task-specific keypoints collected at runtime to represent object-centric local reference frames throughout the workspace.}
\label{fig:movie1}
\end{figure}

\subsection{System Overview}

Our system comprises a 6-DoF uFactory Lite 6 robot equipped with an Intel RealSense D405 stereo camera, a BotaSys SensONE force/torque sensor, and 3D-printed tool mounts for a knife, peeler, and probe (Fig.~\ref{fig:skill_transfer}D).

The inputs to our method are the point cloud collected online and a set of keypoints that can be either provided by the user, automatically extracted or transferred with a learning based approach~\cite{oquabDINOv2LearningRobust}. Using the object point cloud and the keypoints, our representation provides smoothly varying local reference frames at any point in the workspace. Fig.~\ref{fig:overview}A illustrates the workflow, and the full procedure is described in \emph{Materials and Methods}. Different high level controllers can query the local reference frame at a given position and generate local action references for various downstream tasks. These local action references are then tracked by an admittance controller~\cite{scherzingerForwardDynamicsCompliance2017}. The overall workflow is depicted in Fig.~\ref{fig:overview}B.

\subsection{Object-centric Task Representation and Transfer}
\label{sec:task_transfer}

\begin{figure}[]
\centering
\includegraphics[width=1\linewidth]{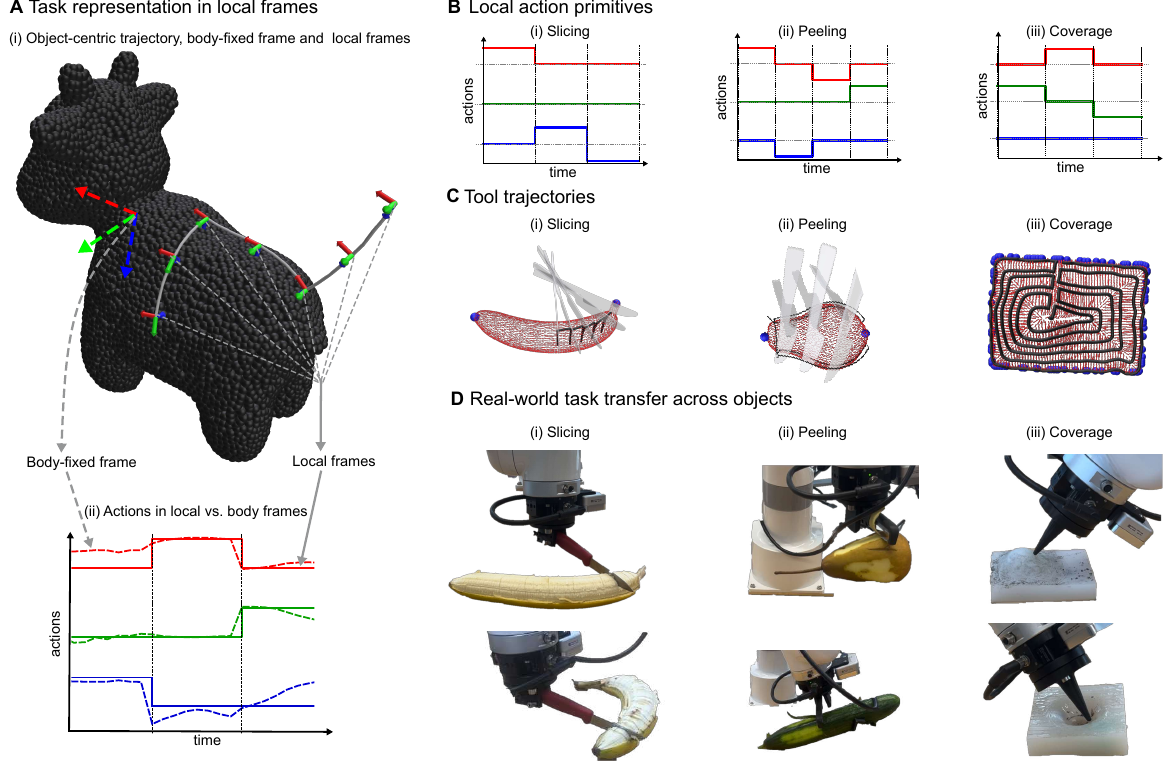}
\caption{\textbf{Shape-invariant task representation and transfer using DOF}. 
(\textbf{A}) (i) An object-centric trajectory (gray) composed of an approaching phase and a surface-following phase. Small reference frames are the object-centric local reference frames represented by DOF, while the large dashed frame denotes a single body-fixed frame. (ii) Gray trajectory is expressed in object-centric local frames (solid lines) and the single body frame (dashed lines). (\textbf{B}) Local action primitives providing shape-invariant task descriptions in object-centric local reference frames. (\textbf{C}) Simulated visualizations showing the keypoints (blue), tool paths (black), and the x-axis of local frames (red). (\textbf{D}) Real-world experiments showing transfer of slicing, peeling, and tactile coverage tasks to novel objects.}
\label{fig:skill_transfer}
\end{figure}

Object-centric local reference frames represented by DOF encodes task-relevant directions across the workspace considering the surface geometry and semantic keypoints. Expressing tasks in these local frames decouples task representation from the object geometry and yields structured action sequences (Fig.~\ref{fig:skill_transfer}A). We refer to these shape-invariant task representations that remain unchanged across variation in object geometry as \emph{local action primitives}. Local action primitives are closed-loop policies in which action transitions are governed by distance to the keypoints and to the surface. We provide local action primitives in Fig.~\ref{fig:skill_transfer}B for three representative object-centric tasks: slicing, peeling and tactile coverage (see \emph{Supplementary Materials} for details).

We transfer tasks represented as local actions to a novel object by computing DOF conditioned on the object's point cloud and by tracking local actions using an admittance controller in local frames. We visualize the tool trajectories for a given object and task in Fig.~\ref{fig:skill_transfer}C and demonstrate real-world transfer across objects in Fig.~\ref{fig:skill_transfer}D.

\begin{figure}[]
\centering
\includegraphics[width=0.9\linewidth]{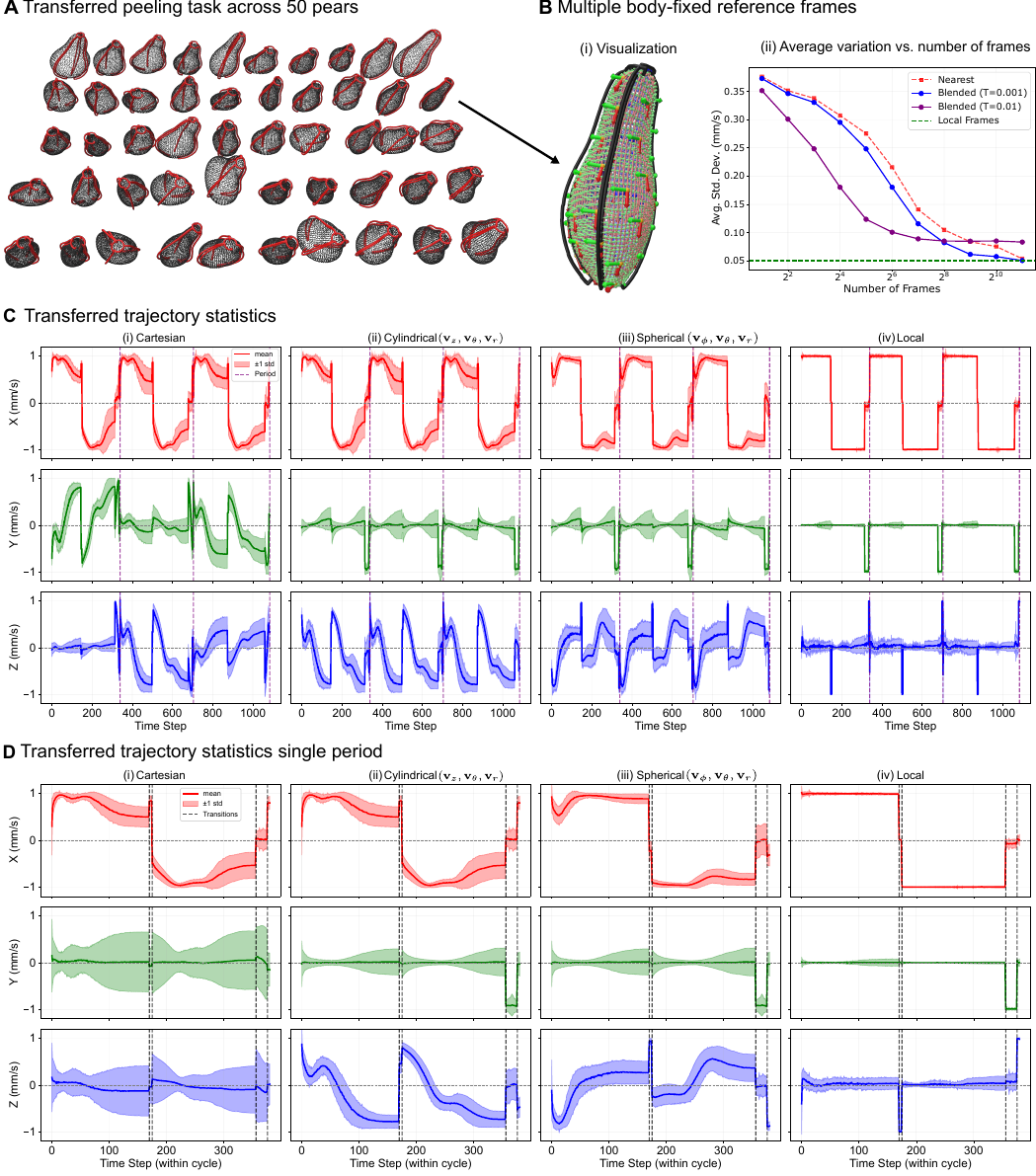}
\caption{{\textbf{Task transfer across objects}. 
(\textbf{A}) Transferred peeling trajectories (red) across 50 pear instances (black). Each trajectory consists of three peeling cycles. (\textbf{B}) Comparison of using multiple discrete body-fixed reference frames versus a field of local reference frames provided by DOF. (i) Visualization of 50 body-fixed reference frames sampled on a pear instance. (ii) Average standard deviation of the transferred action (velocity) trajectories with respect to number of sampled body-fixed reference frames. (\textbf{C} and \textbf{D}) Comparisons of DOF with respect to baselines in terms of the full and period-aligned transferred trajectory statistics, respectively. Action trajectory statistics expressed in (i) Cartesian, (ii) cylindrical, (iii) spherical body-fixed reference frames and (iv) local reference frames provided by DOF.}}
\label{fig:transfer_comparison}
\end{figure}

To evaluate task transfer quantitatively and compare our approach to alternative object-centric representations, we transferred the peeling task to 50 randomly deformed versions of the Pear object (Fig.~\ref{fig:transfer_comparison}A) from the YCB dataset~\cite{
calliYCBObjectModel2015}. We generated deformed object instances by applying random anisotropic scaling, quadratic bending, and twist deformations. For each instance, we first sampled independent scale factors \(s_x, s_y, s_z \sim \mathcal{U}(0.6, 1.4)\) and scaled the coordinates along the corresponding axes. To introduce nonuniform deformations, we then applied quadratic bending along each Cartesian axis. For bending about axis \(j \in \{x,y,z\}\), we sampled a curvature parameter \(c_j \sim \mathcal{U}(-3, 3)\) and displaced the coordinates along the two orthogonal axes \(i \neq j\) according to
\begin{align}
p_i &\leftarrow p_i + c_j\, p_j^2.
\end{align}
In addition, we applied twist deformations around each axis. For twisting about axis \(j\), we sampled a twist strength \(\alpha_j \sim \mathcal{U}(-3, 3)\) and rotated each point around axis \(j\) by an angle
\begin{align}
\theta_j &= \alpha_j d_j,
\end{align}
where \(d_j\) is the coordinate of the point along the twist axis.

For comparisons, we selected four object-centric representations as baselines:
(i) a body-fixed Cartesian frame,
(ii) a body-fixed cylindrical frame,
(iii) a body-fixed spherical frame, and
(iv) multiple body-fixed frames distributed at equal spacing over the object surface (Fig.~\ref{fig:transfer_comparison}B).
A single body-fixed frame is the most common object-centric choice because it provides pose-invariance, but it ignores the geometry of the shape. The cylindrical and spherical variants retain pose invariance and additionally encode axial and radial symmetry, respectively~\cite{tiGeometricOptimalControl2023}. To ensure a fair comparison and take advantage of object symmetry, we aligned each approach with the primary symmetry axis of the object, defined by the line connecting the two keypoints used in our method. We then expressed the actions as linear velocities along the three orthogonal directions induced by each approach.

The multiple body-fixed frames baseline serves two purposes. First, it is a discrete approximation of our approach, since DOF yields a continuous field of local frames over the surface. Second, it represents a common family of methods in the literature, including task-parameterized models~\cite{calinonTutorialTaskparameterizedMovement2016}, oriented keypoints~\cite{gaoKPAM20Feedback2021}, and neural descriptor fields~\cite{simeonovNeuralDescriptorFields2022}. Importantly, our instantiation is an oracle version of these approaches, because it uses exact keypoint correspondences and the task aligned local frames provided by our method. This choice favors the baseline and thus makes the comparison conservative.

As the comparison metric, we use statistics of the transferred action trajectories. An ideal object-centric task representation for transfer should be shape invariant and therefore produce lower variation across shapes. We report the mean and the standard deviation of 50 trajectories that each contain three peeling cycles (Fig.~\ref{fig:transfer_comparison}C). Our representation yields the smallest variation because it decouples object geometry from the task coordinates, that is, it is shape-invariant. It also preserves the three cycle periodic pattern in all coordinates, which shows that the global symmetry of the object is captured while remaining separate from the task representation. In contrast, the other baselines capture only the particular symmetry they encode and only along one or two directions.
To make this effect explicit, we further decompose each trajectory into individual peeling cycles, align them in time, and plot the mean and the standard deviation of the actions for each baseline in Fig.~\ref{fig:transfer_comparison}D.

To show that our method is the continuous version of approaches based on multiple body-fixed reference frames, oriented keypoints, and neural descriptor fields, we measured the average variance of transferred trajectories as a function of the number of frames sampled using Farthest Point Sampling on the point cloud (Fig.~\ref{fig:transfer_comparison}B ii). At each robot position, we express velocities in sampled frames using two variants: (i) nearest sampled frame assignment, and (ii) distance-weighted blending with softmax weights 
$$
w_k=\frac{\exp \left(-d_k / T\right)}{\sum_j \exp \left(-d_j / T\right)},
$$
where $d_k$ is the distance from the robot position to frame $k$ and $T$ is a temperature parameter. As expected, the variation decreases as the number of frames increases, converging to our continuous local frame method.

From an imitation learning perspective, the transferred trajectories can be treated as demonstrations of the peeling task across different object instances. Our representation yields lower variation, directional decoupling, and periodic structure, which should enable better learning performance for statistical learning methods.

\subsection{Integration with Different Control Paradigms}
\label{sec:controller_transfer}
\begin{figure}[]
    \centering
    \includegraphics[width=1.0\linewidth]{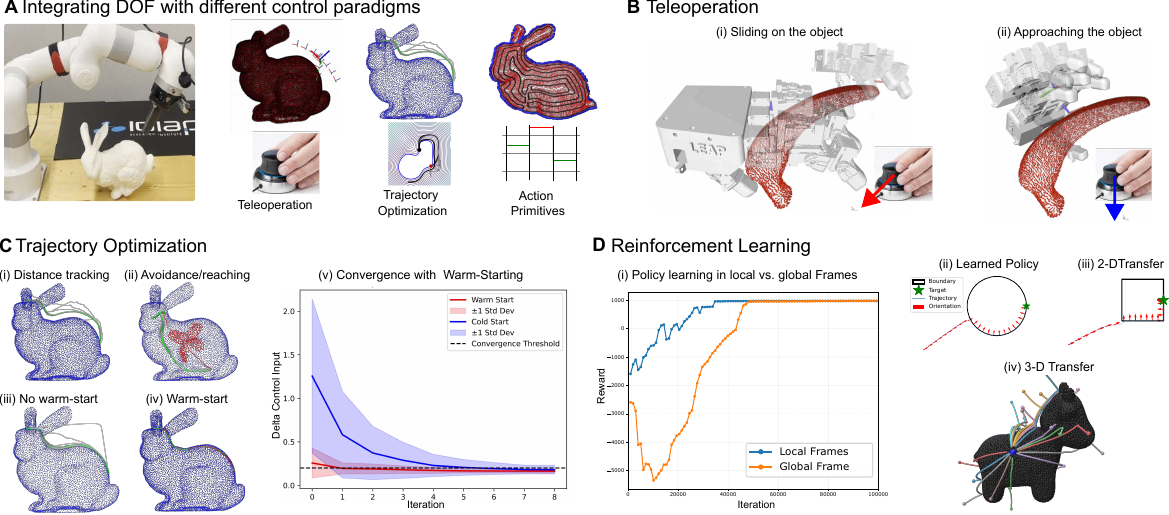}
    \caption{\textbf{Transfer across controllers using DOF.} (\textbf{A}) DOF as a controller agnostic intermediate representation which can be integrated with various reactive and anticipative controllers. (\textbf{B}) Teleoperation using a space mouse and the LEAP hand~\cite{shawLEAPHandLowCost2023}, where the input axes are mapped to local frames. Moving along x-axis (red arrow) slides the tool along the surface, while z-axis (blue arrow) approaches the surface. (\textbf{C}) Trajectory optimization experiments for (i) distance tracking, (ii) target reaching and obstacle avoidance, (iii) reaching without warm-starting and (iv) reaching with warm-starting using the DOF. (v) Norm of the change in the control commands, which is
    used as the convergence criteria for the trajectory optimization, showing the effect of warm-starting. (\textbf{D}) Learning and transferring policies using reinforcement learning in local reference frames. (i) Reward evolution while learning a reaching policy using local and global reference frames. (ii) Learned target reaching and distance tracking policy in local reference frames of a 2-D circle. Zero-shot transfer of the learned policy on the circle to (iii) a 2-D rectangle and (iv) a 3-D point cloud.}
    \label{fig:controller_transfer}
\end{figure}

DOF serves as an intermediate geometric representation that is agnostic to how high-level commands are generated, enabling seamless integration with diverse control paradigms (Fig.~\ref{fig:controller_transfer}A). In addition to the local action primitives presented in the previous section, we integrated DOF in three settings to show it reduces controller complexity by decoupling the action space from object geometry: a teleoperation controller, a planner based on trajectory optimization, and a policy learned with reinforcement learning.

In teleoperation experiments, we used DOF to assist the operator by automatically aligning the tool based on the object's surface and task-specific keypoints. We employed the 3DConnexion Space Mouse, a 6-degree-of-freedom input device, and mapped its control axes to the object-centric local reference frames represented by DOF. Motion along the input device’s x-axis translated to movements that approached or retreated from keypoints while maintaining a constant distance from the surface. Motion along the z-axis controlled approach or retreat relative to the object, and motion along the y-axis preserved the distance to both the keypoints and the surface. Throughout the task, the desired tool orientation—relative to the surface and task direction—was maintained. This setup enabled intuitive, surface-aware teleoperation, as illustrated in Fig.~\ref{fig:controller_transfer}B.

In order to show how DOF can be more broadly applied in complex especially long-horizon robotic manipulation tasks, we integrated it in trajectory optimization for planning. We used DOF to define cost functions and their gradients based on the distance to the object's surface and geodesic distances to surface regions encoded using keypoints. These components enabled gradient-based optimization to compute robot trajectories that both maintain a desired distance from the surface and reach target regions while avoiding obstacles (Fig.~\ref{fig:controller_transfer}C, i-ii). For simplicity, we adopted a batch formulation of the iterative linear quadratic regulator (iLQR), modeling the robot as a velocity-controlled point mass. To further improve convergence, we warm-started the optimization using geodesic shooting (i.e., initializing the trajectory by following the x-axes of local reference frames represented by DOF) (Fig.~\ref{fig:controller_transfer}C, iii-iv). We evaluated this setup across ten different initial conditions and plotted the number of iterations required for convergence (Fig.~\ref{fig:controller_transfer}C, v). Results show that with DOF-based warm-starting, convergence is typically achieved in a single iteration, whereas without warm-starting, at least five to six iterations are required. Qualitative observations of the initial and final trajectories further support this finding. In essence, DOF provides a near-optimal solution for distance tracking, reaching, and surface-based obstacle avoidance purely from geometric information.

We ran proof of concept experiments to integrate DOF with reinforcement learning. First, we trained a two dimensional reaching policy and compared using local reference frames versus a global frame, measuring reward evolution over training (Fig. \ref{fig:controller_transfer}D, i). Even in this simple setting, providing a geometric scaffold through local frames improves learning performance. Next, we considered a two dimensional reaching and distance tracking task on a circle (Fig. \ref{fig:controller_transfer}D, ii). Using the same environment and reward, we trained two policies: a local policy that issues actions in DOF frames and a body-fixed policy that acts in a single frame attached to the object. We then deployed both policies on novel shapes. The local policy learned in DOF frames transfers directly to new two dimensional shapes such as a square and to three dimensional objects such as the Spot point cloud (Fig.~\ref{fig:controller_transfer}D, iii-iv). In contrast, the policy trained in a body-fixed frame does not transfer to other shapes. Additional details on local policy learning are provided in the \emph{Supplementary Materials}.
\subsection{Robustness to Sensing Imperfections}
\label{sec:robustness}
\begin{figure}[]
    \centering
    \includegraphics[width=1.0\linewidth]{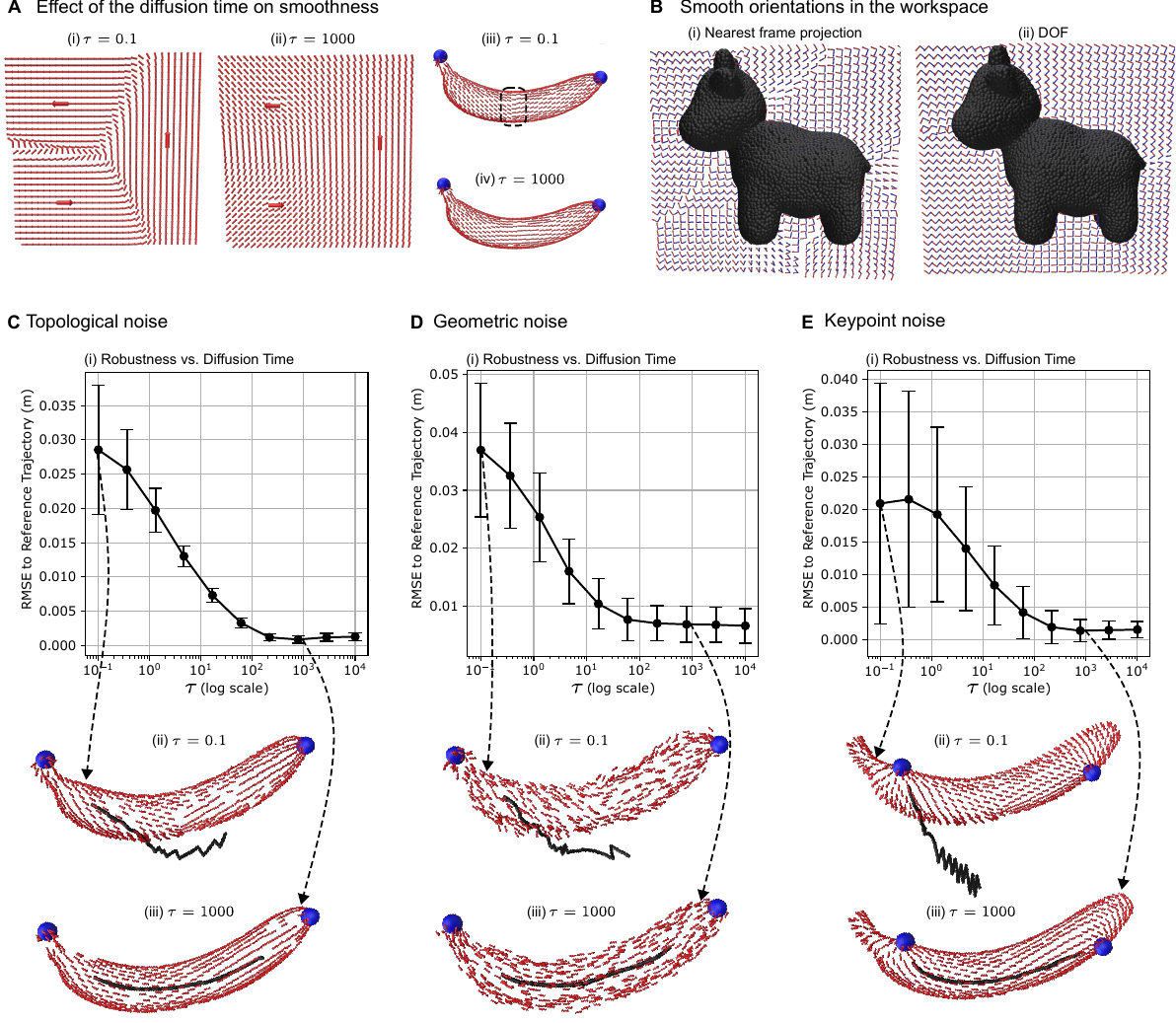}
    \caption{\textbf{Effect of increasing the diffusion time on robustness to noise.} We visualize the local reference frames using their x-axes shown in red. (\textbf{A}) Effect of the diffusion time on the smoothness of the DOF. (\textbf{B}) Comparison of projection to surface orientation field versus the smooth orientation field produced by the DOF. (\textbf{C}, \textbf{D} and \textbf{E}) (i) Error bar plots showing that robustness to topological, geometric and keypoint noise increases with the increased diffusion time. (ii and iii) Two instances with noisy inputs from the experiments visualizing the effect of the short and long diffusion times. Red arrows are the x-axes of the local reference frames, the black curve is the trajectory compared to the reference trajectory and the blue points are the keypoints.}
    \label{fig:robustness}
\end{figure}

In robotic applications, sensor data is often noisy or incomplete, and control or tool positioning may be imprecise. Our representation provides robustness in such scenarios through its inherent smoothness, governed by the \emph{diffusion time} parameter $\tau$. This parameter controls the extent of smoothing in the orientation field: lower values emphasize local geometric details, while higher values produce smoother fields that reflect the global symmetry structure of the shape. It is important to note that diffusion time is only a parameter that controls the smoothness of the field, and is not related to the time required to compute the field. We illustrate the influence of diffusion time on field smoothness in Fig.~\ref{fig:robustness}A, and compare the DOF to a baseline method using projection to the surface orientation field in Fig.~\ref{fig:robustness}B.

While real-world experiments already demonstrate the effectiveness of our method under realistic sensor noise, we further evaluated its robustness under controlled synthetic perturbations. Specifically, we investigated how increasing the diffusion time affects robustness in the presence of various noise types. All experiments were conducted on the Banana point cloud from the YCB dataset~\cite{calliYCBObjectModel2015}.

We performed three experiments, each repeated 50 times with randomly sampled noise:
\begin{itemize}
    \item \textbf{Topological noise:} To simulate occlusions and limited viewpoints, we removed half of the point cloud and introduced ten randomly placed holes with a 5\,mm radius, disrupting surface connectivity.
    \item \textbf{Geometric noise:} Gaussian noise with standard deviation \( \sigma = 3\,\text{mm} \) was added to the 3D point positions.
    \item \textbf{Keypoint noise:} Gaussian noise with standard deviation \( \sigma = 20\,\text{mm} \) was applied to both keypoints, which were then projected back onto the nearest point cloud vertices.
\end{itemize}

To quantify robustness, we computed the root mean square error (RMSE) between each resulting trajectory and a reference trajectory. The reference was generated using noise-free inputs, diffusion time \( \tau = 1000 \), and tracking the local x-direction of the DOF field.

As expected from the diffusion equation which suppresses high frequency noise, and as confirmed empirically in our real-world experiments, increasing diffusion time consistently produces smoother and more robust orientation fields across all tested perturbations (Fig.~\ref{fig:robustness}C–E). However, when depth returns are severely degraded or absent (for example, when using stereo camera on transparent objects), smoothing alone cannot recover missing geometry. In these scenarios, our method requires to use a partial reconstruction/completion method or additional sensors providing better depth data. Additional experiments in clutter and with occlusions are provided in the \emph{Supplementary Materials}.

\subsection{Multi-Object Scenes and Compositionality}
\label{sec:multi_object_scenes}
\begin{figure}[]
    \centering
    \includegraphics[width=1.0\linewidth]{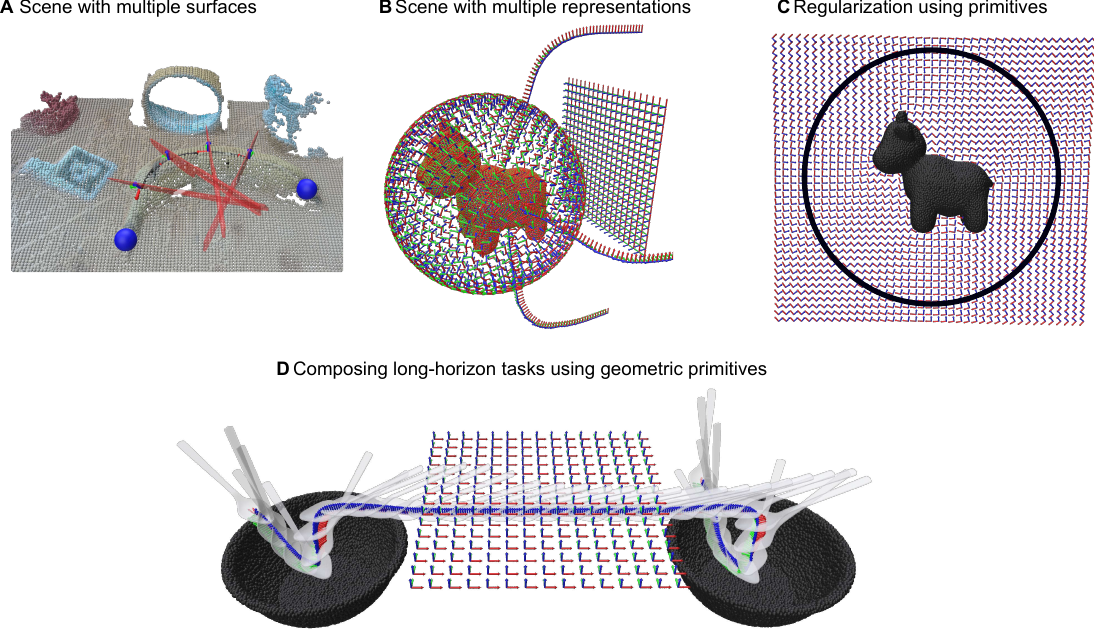}
    \caption{\textbf{DOF constructed from point clouds and primitive shapes.} (\textbf{A}) Slicing on a banana in a cluttered scene with multiple objects. (\textbf{B}) Object point cloud surrounded with an enclosing sphere and a plane designating a wall in the workspace. Trajectories starting from different surface points and following local z-directions to visualize smooth variation of local reference frames across the  workspace. (\textbf{C}) Cross section of the diffused orientation field around the object, regularized with an enclosing sphere. (\textbf{D}) A scoop-lift-carry-pour task using two bowls from the YCB dataset~\cite{calliYCBObjectModel2015}, together with lines and a plane to impose task-specific constraints.}
    \label{fig:primtives}
\end{figure}

DOF is agnostic to the surface representation and the number of objects in the scene. This enables its application in cluttered environments containing multiple surfaces with diverse representations (Fig.~\ref{fig:primtives}A-B). Interestingly, cluttered workspaces can improve the computational efficiency of DOF due to the specific method used for its computation; such environments tend to result in mostly enclosed regions rather than unbounded ones. For instance, in the setup shown in Fig.~\ref{fig:primtives}B, introducing an enclosing sphere reduces the computational cost by $\approx 1.5\times$. We discuss this further in the Materials and Methods section.

DOF supports any surface representation, provided that closest-point queries can be computed. In our experiments, we focused on primitive shapes defined by analytical distance functions. In simple cases, primitives can approximate basic surfaces (e.g., walls, planes, or workspace boundaries) or compose more complex environments through combinations of primitives (e.g., spheres and capsules for obstacle avoidance). More interestingly, primitives can encode task constraints directly into the DOF representation. For example, enclosing an object point cloud with a sphere can regularize the orientation field far from the object (Fig.~\ref{fig:primtives}C).

We modeled a proof-of-concept long-horizon scooping task with multiple objects using DOF (Fig.~\ref{fig:primtives}D). Task constraints such as keeping the tool horizontal to prevent spillage and specifying the lifting direction were encoded directly as simple geometric primitives: a plane to enforce level orientation and a line to define upward/downward motion. This construction introduces no extra parameters for combining DOF and requires no behavior tuning. Because DOF solves for the smoothest orientation field consistent with these constraints across the workspace, the controller is naturally robust to different initial conditions and moderate perturbations.

\subsection{Comparisons}
We qualitatively compared our method against five commonly used representations in robotics:
\begin{itemize}
    \item \textbf{Primitives} represent objects using simple geometric shapes (e.g., spheres, cylinders, boxes) with analytical distance functions. They are computationally efficient and well-suited for collision avoidance, but lack the geometric detail required for tasks involving contact or fine surface interaction. 

    \item \textbf{Surface Parameterizations} map points on a surface to coordinates in $\mathbb{R}^2$. They are limited to the surface and inherently introduce distortion due to curvature.


    \item \textbf{Signed Distance Fields (SDFs)~\cite{fengHeatMethodGeneralized}} implicitly represent surfaces by encoding the distance to the nearest surface point. Although they support level set representations and are useful in 3D space, they lack geodesic information.

    \item \textbf{Neural Descriptor Fields (NDFs)~\cite{simeonovNeuralDescriptorFields2022}} use neural networks to learn smooth implicit surface representations from data. While expressive and capable of handling various tasks, existing approaches are typically limited to scalar fields and require offline training.

\end{itemize}
We compare these representations across key properties relevant to object-centric tasks in Table~\ref{tab:representation_comparison}. DOF’s primary advantage over alternative representations is its ability to encode both geodesic and Euclidean information, enabling geometry-aware interactions on and across level sets.
\begin{table}[ht]
\centering
\begin{tabular}{lccccc}
\hline
 & \textbf{Primitives} & \textbf{SDFs} & \textbf{NDFs} & \textbf{Surface Param.} & \textbf{DOFs} \\
\hline
Geodesic-aware        & \cmark & \xmark & \xmark & \cmark & \cmark \\
Extends to workspace  & \cmark & \cmark & \cmark & \xmark & \cmark \\
No training required  & \cmark & \cmark & \xmark & \cmark & \cmark \\
Online updates        & \cmark & \cmark & \xmark & \xmark & \cmark \\
Exact distances       & \cmark & \cmark & \xmark & \cmark & \xmark \\
Contact-rich tasks    & \xmark & \cmark & \cmark & \cmark & \cmark \\
\hline
\end{tabular}
\caption{\textbf{Comparison of object representations across key properties relevant to manipulation and planning.}}
\label{tab:representation_comparison}
\end{table}
\subsection{Discussion}
\label{sec:discussion}
The proposed Diffused Orientation Fields (DOF) approach smoothly represents object-centric local reference frames across the workspace, conditioned on object point clouds and keypoints collected online. Since DOF encodes surface's geometric structure, it enables shape-invariant task representations and their transfer across curved objects. We demonstrated the transfer capabilities of DOF in real-world experiments with three distinct tasks across six different objects (Fig.~\ref{fig:skill_transfer}). We assessed the robustness of DOF under noise in keypoint extraction, point cloud data (Fig.~\ref{fig:robustness}). These experiments highlight the inherent robustness of our method, owing to the smoothing effect of the diffusion PDE, making it well-suited for robotics scenarios characterized by noisy and incomplete sensory data. Our results show that DOF scales out-of-the-box to complex scenes including multiple objects and diverse surface representations (Fig.~\ref{fig:primtives}) and its computational efficiency (Table~\ref{tab:performance}) enables online updates to the field in changing scenes.

We compared DOF qualitatively and quantitatively to commonly used object-centric representations in robotics (Fig.~\ref{fig:transfer_comparison} and Table~\ref{tab:representation_comparison}). DOF seamlessly combines the gradient of the signed distance field (SDF) to the surface with the gradients of geodesic distances to keypoints, yielding a single smooth field. This integration allows us to reason about directions pointing to keypoints on the level sets and directions pointing towards the surface. In contrast to DOF, learning-based approaches such as neural descriptor fields rely on data-driven learning to implicitly capture structures, offering greater expressiveness but at the cost of requiring training data. DOF takes a different stance: it treats keypoints as encoding the inductive biases relevant to the task and propagates this structure across the workspace using a diffusion process. This reframes task transfer across objects as the simpler problem of transferring keypoints. This trade-off is often advantageous since keypoints can be extracted using simple perception pipelines (e.g., boundary detection), transferred through foundational models as shown in prior work, or manually annotated when needed, given that they form a sparse and interpretable set.

Given that DOF is designed for manipulation tasks that can include dynamic objects, we evaluated its suitability for online computation (Table~\ref{tab:performance}). We observed that the most expensive operation is the construction of the Laplacian operator from the object's point cloud, computed during preprocessing. At runtime, this operator can be reused even under isometric deformations encapsulating rigid body transformations and bending without stretching. Although we do not evaluate this explicitly in the current paper, this property makes DOF applicable to deformable objects such as garments or cables. However, we note that interactions with soft objects alter the underlying geodesic structure, and therefore the precomputed Laplacian may need to be updated to reflect significant deformations.

A core limitation of using the Walk on Spheres (WoS) method is its restriction to solving Laplace's equation, which yields the smoothest possible vector field given the boundary conditions. While this is beneficial for robustness, it also limits expressiveness in scenarios where more localized or structured behavior is desired. One promising direction to address this limitation, while still leveraging the computational advantages of WoS, is to solve the screened Poisson equation instead of Laplace's equation~\cite{sawhneyMonteCarloGeometry}. The screening term introduces a tunable decay, enabling better control over the spatial extent and locality of the resulting field.

Another practical limitation of our approach is the need to predefine the tool center point (TCP) before the execution. This requires precise calibration of the tool. In real-world scenarios, especially in contact-rich tasks, the actual contact point may shift dynamically along the tool during interaction. We showed in experiments that, due to smoothness, DOF provides robustness against positioning errors of the robot. Nevertheless, a more robust alternative would involve estimating the effective TCP online by combining vision and force/torque feedback. DOF queries could then be performed at these dynamically inferred contact points, improving reliability during complex manipulation.

Complex object-centric tasks can be composed from local actions expressed in local frames through planning or learning. For planning, trajectory optimization can exploit gradients provided by DOF for costs and constraints that follow the surface and keypoints (Fig. 5C). This can enable planning sequences that alternate between approach, contact, sliding, and retreat while remaining consistent with the object geometry. For policy learning, expressing observations, actions, rewards, and demonstrations in local frames provides pose and shape-invariance (Figs. 4 and 5D). This in turn can allow demonstrations from different objects to be combined into a unified policy, since their local descriptions become comparable under the DOF structure, and supports transferring a policy learned on one object to another. Moreover, our entire pipeline is differentiable with respect to both the task keypoints and the diffusion time parameter~\cite{sharpDiffusionNetDiscretizationAgnostic2022, millerDifferentialWalkSpheres2024}. Therefore a promising future extension is to integrate DOF into learning and optimization based frameworks where these parameters are optimized jointly with task performance.
\section{Materials and Methods}
\label{sec:background}
We first present the necessary background, then we formulate our problem and finally we present our method for computing the diffused orientation fields.

\subsection{Background}
\subsubsection{Smooth Fields Via Diffusion Processes}
Consider a scalar field $u : \Omega \rightarrow \mathbb{R}$. We can measure the smoothness of the field using the Dirichlet energy functional $E[u]$ penalizing large gradients
\begin{align}
    E[u]=\int_{\Omega}\|\nabla u\|^2 d \boldsymbol{x} = \int_{\Omega} (\nabla u \cdot \nabla u) d \boldsymbol{x}.
   \label{eq:dirichlet_energy}
\end{align}
Accordingly, finding the smoothest field over the domain subjected to the fixed values, i.e. Dirichlet boundary conditions, is given by the variational optimization problem:
\begin{align}
    \min _{u\in \mathcal{A}} E[u],
    \label{eq:variational_optimization}
\end{align}
constrained to the admissable set
\begin{equation}
\mathcal{A}=\left\{u \in H^1(\Omega):\left.u\right|_{\partial \Omega}=g\right\},
\end{equation}
where $H^1(\Omega)$ is the space of functions with square-integrable derivatives required for defining Dirichlet energy and $\left. u\right|_{\partial \Omega}=g$ are the boundary conditions where $\partial \Omega$ denotes the boundary. The solution to the variational problem in~\eqref{eq:variational_optimization} is given by the partial diferential equation called Laplace's equation
\begin{equation}
    \Delta u=0 \text { in } \Omega,
    \label{eq:laplaces_equation}
\end{equation}
subjected to the boundary conditions:
\begin{equation}
\begin{array}{ll}
u=g & \text { on } \partial \Omega, \\
\nabla u \cdot \boldsymbol{n}=0 & \text { on } \partial \Omega.
\end{array}
\end{equation}
where the differential operator $\Delta$ is called the Laplacian. The solution to Laplace's equation provides the smoothest possible interpolation of Dirichlet boundary conditions across the interior of the domain $\Omega$. Alternatively, one can control the smoothing using the gradient flow of Dirichlet energy which models a time-dependent field $u(\bm{x}, t)$, starting from the initial condition $u(\bm{x}, 0)$ and gets smoother over time:
\begin{equation}
\dot{u}=\Delta u \quad \text{in} \quad \Omega.
\label{eq:diffusion}
\end{equation}
This equation is the diffusion PDE.

\subsubsection{Discretized Laplacian on Curved Surfaces}
\label{subsec:laplacian}
Surfaces are two-dimensional manifolds embedded in 3-D Euclidean space i.e., topological spaces that locally look like Euclidean space but not necessarily globally. Unlike the flat geometry of Euclidean space, curved manifolds do not admit a global coordinate frame to measure distances, compute angles between tangent vectors, or define differential operators such as the Laplacian. Laplace-Beltrami operator $\Delta_{\mathcal{M}}$ generalizes the Laplacian from Euclidean space to curved spaces but for conciseness, we will refer to Laplace-Beltrami operator as the Laplacian. In practice the Laplacian is only available in closed form for a limited set of highly structured manifolds such as cylinders or spheres. Accordingly, for an arbitrary surface, it is necessary to approximate the Laplacian using a discrete representation of the surface.

For a discrete surface representation such as a mesh, a point cloud, or a grid with $\npoints$ vertices, the discretized Laplacian $\bm{L} \in \mathbb{R}^{\npoints \times \npoints}$ can be approximated using the sparse matrices 
\begin{equation}{\label{eq:discrete_Laplacian}}
    \bm{L}=\bm{M}^{-1}\bm{C}, 
\end{equation}
where $\bm{M}$ is the diagonal mass matrix and $\bm{C}$ is a sparse symmetric matrix called the weak Laplacian. The entries of $\bm{M}$ correspond to the area around each vertex and the entries of $\bm{C}$ are determined by the connectivity of the points on the local tangent space and the distance between the connected points. In particular, we use the approach proposed by Sharp~\emph{et al.}~\cite{sharpLaplacianNonmanifoldTriangle2020} for computing the Laplacian of the point clouds.

\subsubsection{Walk on Spheres}
\label{subsec:walk_on_spheres}
Walk on Spheres (WoS) is a grid-free Monte Carlo method for solving Laplace's equation~\eqref{eq:laplaces_equation} in Euclidean spaces. A function which satisfies Laplace's equation is called an harmonic function and it satisfies the \emph{mean-value property} i.e., $u(\bm{x})=\mathbb{E}[u(\bm{X})]$ where $\bm{X}$ is a random point on a sphere centered at $\bm{x}$ with radius $r$. Accordingly, we can simulate a random process where a particle starts $\bm{x}$ and jumps to a new position randomly chosen on a sphere centered at $\bm{x}$. This process repeats until the particle reaches the boundary, i.e., surface of the object and the function value is given by the value at the boundary point the particle landed. We refer the readers to~\cite{sawhneyMonteCarloGeometry} for a thorough analysis on WoS.

\subsection{Problem Formulation}
As we showed in Fig.~\ref{fig:skill_transfer}, object-centric tasks can be expressed as simple action sequences in object-centric local reference frames. While these action sequences are simple to express in a fixed frame for flat surfaces, curved objects lack a global reference frame, and the task-relevant directions vary across the surface. To address this, we compute local reference frames that smoothly vary by considering the object's geometry to represent object-centric tasks. 


At each point $\bm{x} \in \Omega$ in the robot's workspace, we represent a local reference frame by its orientation relative to the world frame. Collectively, these orientations form an orientation field $\bm{u}(\bm{x}): \Omega \to SO(3)$, assigning a local reference frame to every point in the workspace. We quantify the smoothness of this field by the Dirichlet energy, as defined in Equation~\eqref{eq:variational_optimization}. Then, we pose the problem of computing a smooth orientation field $\bm{u}(\bm{x})$ as a variational optimization problem, whose solution corresponds to Laplace's equation~\eqref{eq:laplaces_equation}, or to the diffusion equation~\eqref{eq:diffusion} for controlled smoothness. These PDEs enforce spatial smoothness but require initial and boundary conditions for well-posedness and uniqueness. We use a sparse set of keypoints as the initial or boundary conditions to encode task-relevant directional cues that serve as anchors for the orientation field. 

In the following subsections, we describe how we combine the diffusion and Laplace's equations with object point clouds and keypoints to generate orientation fields. First, we present computing orientation fields on object surfaces using surface diffusion conditioned on the keypoints. Then, we show propagating these orientations from the object surface to the robot's workspace by solving Laplace's equation~\eqref{eq:laplaces_equation} in the surrounding Euclidean space. Lastly, we describe the computation of the orientation fields in the scenes using multiple surfaces with different representations such as a point cloud, a mesh, or a geometric primitive.

\subsection{Diffused Orientation Fields on Surfaces}
    We measure the object surfaces at runtime as point clouds $\mathcal{P}$ composed of $\npoints$ points
    \begin{equation}
        \mathcal{P}:=\left\{\begin{array}{l|l}
        \left(\boldsymbol{x}_i\right) & \begin{array}{l}
        \boldsymbol{x}_i \in \mathbb{R}^3 \\
        \text { for } i=1, \ldots, \npoints
        \end{array}
        \end{array}\right\}
    \end{equation}
    where $\bm{x}_i$ is the position of the $i$-th surface point in Euclidean space. In order to solve the diffusion or Laplace's equation on the point cloud we compute the weak Laplacian $\bm{C}$ and the corresponding mass matrix $\bm{M}$.

    \subsubsection{Orientation Fields from Keypoints}
    \label{subsec:scalar_diffusion}
    The second input of our method is a set of keypoints on the surface~$\mathbf{p}=\left\{\bm{x}_i\right\}_{i=1}^N \subset \mathcal{P}$, which can be either specified by the user, extracted by the perception system or transferred using a vision encoder (see \emph{Supplementary Material}). We use the keypoints for aligning the orientation field with task-relevant directions. We classify each keypoint either as a \emph{sink} acting as an attractor or a \emph{source} acting as a repulsor. Our strategy for closed surfaces is to use one source and one sink placed as two poles (see Fig.~\ref{fig:skill_transfer}C, i-ii). This produces a consistent direction field defining longitudes and latitudes for the object and it is easy to extract and transfer using perception modules. If the object has a boundary, we consider all the boundary points as sources (see Fig.~\ref{fig:skill_transfer}C, iii). In more complex scenarios where we have arbitrary target and/or obstacle regions on the surface, it is possible to set sources and sinks accordingly (see Fig.~\ref{fig:controller_transfer}B, ii). Next, we use the keypoints to set the initial condition $\bm{u}_0$, where the values are assigned using an indicator function, setting entries to $+1$ for source keypoints, $-1$ for sink keypoints, and $0$ elsewhere:
    \begin{equation}
        \bm{u}_0 = 
        \begin{cases}
            +1 & \text{if } \boldsymbol{x} = \mathbf{p}_{\text{source}} \in \mathcal{P}, \\
            -1 & \text{if } \boldsymbol{x} = \mathbf{p}_{\text{sink}} \in \mathcal{P}, \\
            0 & \text{otherwise}.
        \end{cases}
        \label{eq:sources_sinks}
    \end{equation}
    Here we solve the diffusion equation~\eqref{eq:diffusion} to propagate the information from the keypoints across the surface. In the discrete setting, one can solve the diffusion equation by integrating using the implicit time-stepping scheme
    \begin{equation}
        \bm{u}_{\tau}=(\bm{M}-\tau \bm{C})^{-1} \bm{M} \bm{u}_0,
        \label{eq:implicit_time_stepping}
    \end{equation}
    where $\tau$ is the diffusion time, $\bm{u}_0$ is the initial condition and $\bm{u}_{\tau}$ is the diffused field arrays indexed by the points in $\mathcal{P}$. Diffusion time $\tau$ is a hyperparameter that controls the smoothness of the field and is unrelated to wall clock time or to the physical time $t$ used for robot control.
    In other words, large values for the diffusion time parameter emphasize global structure and symmetries of the object\footnote{
    By denoting $\{\phi_k\}_{k=0}^{\infty}$ the Laplace eigenfunctions on the domain with eigenvalues $\{\lambda_k\}_{k=0}^{\infty}$, the diffusion solution can be described with $u(\tau,x) = \sum_{k=0}^{\infty} e^{-\lambda_k \tau}\, c_k\, \phi_k(x)$, where $c_k$ are the coefficients for the initial field $u(x,0)$ in the Laplacian basis.  As $\tau$ increases, high frequency modes (larger $\lambda_k$) are suppressed more quickly by the factor $e^{-\lambda_k \tau}$, so that the result is dominated by low-order modes. On shapes with bilateral or near rotational symmetry, these low-order modes respect the symmetry group, therefore the field aligns with global symmetry as $\tau$ grows.}.

    To remove dependence on spatial scale, by following common practice~\cite{craneGeodesicsHeatNew2013}, we normalize the parameter by the squared mean neighbor spacing $h^2$ of the point cloud,
    \begin{equation}
    \tau \leftarrow \frac{\tau}{h^{2}}.
    \label{eq:diffusion_time}
    \end{equation}

    Solving the linear system in \eqref{eq:implicit_time_stepping} is very efficient since it reduces to sparse matrix multiplication if one pre-computes the $\bm{C}$, $\bm{M}$ and matrix factorization after obtaining the point cloud (see Table~\ref{tab:performance}). Next, we compute the gradient of the diffused field $\nabla_{\bm{x}} \bm{u}_\tau$ to obtain a smooth direction field on the surface. We visualize these direction fields using red arrows in Figure~\ref{fig:overview}B. The direction vector at a given point $\bm{x}_i$ lies on the tangent plane of the surface at that point $\mathcal{T}_{\bm{x}_i}\mathcal{P}$. The tangent plane is spanned by two perpendicular vectors $\mathbf{u}$ and $\mathbf{v}$. We use the gradient of the diffused field as the first tangent vector $\mathbf{u} = \nabla_{\bm{x}} \bm{u}_\tau$. Then one can find the perpendicular tangent vector using the local surface normal $\mathbf{n}$ and the cross product $\mathbf{v} = \mathbf{n}\times \mathbf{u}$. These three unit vectors are by construction orthogonal to each other and they define both a local reference frame and a proper rotation matrix with positive determinant if they are sequenced in the correct order $\mathbf{u}$, $\mathbf{v}$ and $\mathbf{n}$.

  We prefer to use the diffusion equation~\eqref{eq:diffusion} primarily because it allows us to control the smoothness of the orientation field by adjusting the diffusion time $\tau$. This is a critical aspect since using short-time diffusion results in gradient of the diffused field $\mathbf{u}$ being parallel to the gradient of the geodesic distance (see~\cite{craneGeodesicsHeatNew2013}). In contrast, using long-time diffusion extracts the global symmetry of the objects (see Fig.~\ref{fig:robustness}A, ii and \cite{lipmanBiharmonicDistance2010}). Both of these properties can be useful for different tasks and can be adjusted by a single parameter: the diffusion time $\tau$. Secondly, using the diffusion equation let us set the keypoints as initial conditions allowing us to use pre-factorized matrices when solving~\eqref{eq:implicit_time_stepping}. On the contrary, in Laplace's equation, we need to use boundary conditions altering the discrete Laplacian hence disabling the pre-factorization.

    \subsection{Workspace Diffusion}


    At this stage, we assume that the orientation field has already been computed on the surfaces using either scalar or orientation diffusion depending on the task. Now we extend these fields from surfaces to arbitrary points in the workspace. While one could apply the same orientation diffusion procedure by discretizing the workspace and computing its discrete Laplacian, we instead adopt a Monte Carlo method called \emph{Walk on Spheres} (WoS). WoS avoids the need for discretization, making it better suited for complex, dynamic scenes and generalizable to any surface representation that supports closest point queries.

    The WoS algorithm depends on efficient closest point queries. To enable this, we construct a k-d tree from the point cloud with time complexity $O(n \log n)$ during preprocessing. Each query during runtime then has average complexity $O(\log n)$ and can be efficiently parallelized on the GPU. Unlike the Laplacian-based methods, WoS does not compute the whole diffused field but computes the values of the diffused field in parallel at the query points. In each WoS query, Monte Carlo samples start from the query point in the workspace $\bm{x}_q\in \Omega$ and converge to boundary points on surfaces $\mathbf{x}_i \in \mathcal{P}$. To compute the value of the diffused orientation field at the query point, we need to average the orientations associated with the boundary points.

    For averaging, we represent the retrieved orientations at points $\bm{x}_i$ as a set of unit quaternions  
    $\mathcal{Q} = \{ \mathbf{q}_1, \mathbf{q}_2, \dots, \mathbf{q}_n \}$  
    and employ the method of Markley et al.~\cite{markleyAveragingQuaternions2007}, which solves the following optimization problem:
    \begin{equation}
        \overline{\mathbf{q}} = \arg \max_{\mathbf{q} \in \mathbb{S}^3} \mathbf{q}^\top \bm{M} \mathbf{q},
        \label{eq:quaternion_averaging}
    \end{equation}
    where the matrix $\bm{M}$ is constructed using the outer product of the unit quaternions:
    \begin{equation}
        \bm{M} \triangleq \sum_{\mathbf{q} \in \mathcal{Q}} \mathbf{q} \mathbf{q}^\top.
    \end{equation}
    Note that~\eqref{eq:quaternion_averaging} is equivalent to maximizing the Rayleigh quotient and its solution is given by the eigenvector of $\bm{M}$ with the largest eigenvalue
    \begin{equation}
        \bm{M} \overline{\mathbf{q}}=\lambda_{\max } \overline{\mathbf{q}}.
    \end{equation}
    We provide a conceptually simple approximation for computing DOF using vector diffusion and orthonormalization in the \emph{Supplementary Materials}.

    \subsubsection{Scenes with Multiple Surface Representations}
    In addition to unknown objects collected at runtime as point clouds, complex scenes may include other surfaces represented as meshes or approximated using geometric primitives such as lines, planes, or spheres. By combining the closest point queries for each surface in the scene we can compute the resulting DOF in the workspace considering all the surfaces together.

    For mesh-based surfaces, we apply the same procedure as for point clouds: we first compute the diffused orientation field on the surface using the discrete Laplacian defined on the mesh, and then construct a boundary volume hierarchy (BVH) for efficient closest point queries. For primitive shapes, we instead use analytical solutions: we compute the surface diffusion using the closed-form Laplacian of the primitive and perform closest point queries using the primitives' analytical distance functions. We provide examples of complex scenes with multiple objects and representations in Figure~\ref{fig:primtives}.

\clearpage 

%
\bibliography{ref} 

@inproceedings{amanhoudDynamicalSystemApproach2019,
  title = {A {{Dynamical System Approach}} to {{Motion}} and {{Force Generation}} in {{Contact Tasks}}},
  booktitle = {Proc. {{Robotics}}: {{Science}} and {{Systems}} ({{RSS}})},
  author = {Amanhoud, Walid and Khoramshahi, Mahdi and Billard, Aude},
  year = 2019,
  month = jun,
  publisher = {{Proc. Robotics: Science and Systems (RSS)}},
  doi = {10.15607/RSS.2019.XV.021},
  isbn = {978-0-9923747-5-4}
}

@article{calinonTutorialTaskparameterizedMovement2016,
  title = {A Tutorial on Task-Parameterized Movement Learning and Retrieval},
  author = {Calinon, Sylvain},
  year = 2016,
  month = jan,
  journal = {Intelligent Service Robotics},
  volume = {9},
  number = {1},
  pages = {1--29},
  issn = {1861-2784},
  doi = {10.1007/s11370-015-0187-9}
}

@inproceedings{calliYCBObjectModel2015,
  title = {The {{YCB}} Object and {{Model}} Set: {{Towards}} Common Benchmarks for Manipulation Research},
  shorttitle = {The {{YCB}} Object and {{Model}} Set},
  booktitle = {2015 {{International Conference}} on {{Advanced Robotics}} ({{ICAR}})},
  author = {Calli, Berk and Singh, Arjun and Walsman, Aaron and Srinivasa, Siddhartha and Abbeel, Pieter and Dollar, Aaron M.},
  year = 2015,
  month = jul,
  pages = {510--517},
  doi = {10.1109/ICAR.2015.7251504}
}

@misc{chiDiffusionPolicyVisuomotor2023,
  title = {Diffusion {{Policy}}: {{Visuomotor Policy Learning}} via {{Action Diffusion}}},
  shorttitle = {Diffusion {{Policy}}},
  author = {Chi, Cheng and Feng, Siyuan and Du, Yilun and Xu, Zhenjia and Cousineau, Eric and Burchfiel, Benjamin and Song, Shuran},
  year = 2023,
  month = mar,
  number = {arXiv:2303.04137},
  eprint = {2303.04137},
  primaryclass = {cs},
  publisher = {arXiv},
  doi = {10.48550/arXiv.2303.04137},
  archiveprefix = {arXiv}
}

@article{craneGeodesicsHeatNew2013,
  title = {Geodesics in Heat: {{A}} New Approach to Computing Distance Based on Heat Flow},
  shorttitle = {Geodesics in Heat},
  author = {Crane, Keenan and Weischedel, Clarisse and Wardetzky, Max},
  year = 2013,
  month = oct,
  journal = {ACM Transactions on Graphics},
  volume = {32},
  number = {5},
  pages = {152:1--152:11},
  issn = {0730-0301},
  doi = {10.1145/2516971.2516977}
}

@article{craneRobustFairingConformal2013,
  title = {Robust Fairing via Conformal Curvature Flow},
  author = {Crane, Keenan and Pinkall, Ulrich and Schr{\"o}der, Peter},
  year = 2013,
  month = jul,
  journal = {ACM Transactions on Graphics},
  volume = {32},
  number = {4},
  pages = {1--10},
  issn = {0730-0301, 1557-7368},
  doi = {10.1145/2461912.2461986}
}

@misc{defariasGraspTransferDeformable2022,
  title = {Grasp {{Transfer}} for {{Deformable Objects}} by {{Functional Map Correspondence}}},
  author = {{de Farias}, Cristiana and Tamadazte, Brahim and Stolkin, Rustam and Marturi, Naresh},
  year = 2022,
  month = mar,
  number = {arXiv:2203.00776},
  eprint = {2203.00776},
  primaryclass = {cs},
  publisher = {arXiv},
  archiveprefix = {arXiv}
}

@article{dyckImpedanceControlArbitrary2022,
  title = {Impedance {{Control}} on {{Arbitrary Surfaces}} for {{Ultrasound Scanning Using Discrete Differential Geometry}}},
  author = {Dyck, Michael and Sachtler, Arne and Klodmann, Julian and {Albu-Schaffer}, Alin},
  year = 2022,
  month = jul,
  journal = {IEEE Robotics and Automation Letters},
  volume = {7},
  number = {3},
  pages = {7738--7746},
  issn = {2377-3766, 2377-3774},
  doi = {10.1109/LRA.2022.3184800},
  copyright = {https://ieeexplore.ieee.org/Xplorehelp/downloads/license-information/IEEE.html}
}

@misc{fariasGeometricallyAwareOneShotSkill2025,
  title = {Geometrically-{{Aware One-Shot Skill Transfer}} of {{Category-Level Objects}}},
  author = {de Farias, Cristiana and Figueredo, Luis and Laha, Riddhiman and Adjigble, Maxime and Tamadazte, Brahim and Stolkin, Rustam and Haddadin, Sami and Marturi, Naresh},
  year = 2025,
  month = mar,
  number = {arXiv:2503.15371},
  eprint = {2503.15371},
  primaryclass = {cs},
  publisher = {arXiv},
  doi = {10.48550/arXiv.2503.15371},
  archiveprefix = {arXiv}
}

@article{fengHeatMethodGeneralized,
  title = {A {{Heat Method}} for {{Generalized Signed Distance}}},
  author = {Feng, Nicole and Crane, Keenan},
  volume = {43},
  number = {4}
}

@inproceedings{gaoBiKVILKeypointsbasedVisual2024,
  title = {Bi-{{KVIL}}: {{Keypoints-based Visual Imitation Learning}} of {{Bimanual Manipulation Tasks}}},
  shorttitle = {Bi-{{KVIL}}},
  booktitle = {2024 {{IEEE International Conference}} on {{Robotics}} and {{Automation}} ({{ICRA}})},
  author = {Gao, Jianfeng and Jin, Xiaoshu and Krebs, Franziska and Jaquier, No{\'e}mie and Asfour, Tamim},
  year = 2024,
  month = may,
  pages = {16850--16857},
  publisher = {IEEE},
  address = {Yokohama, Japan},
  doi = {10.1109/ICRA57147.2024.10610763},
  copyright = {https://doi.org/10.15223/policy-029},
  isbn = {979-8-3503-8457-4}
}

@article{gaoKPAM20Feedback2021,
  title = {{{kPAM}} 2.0: {{Feedback Control}} for {{Category-Level Robotic Manipulation}}},
  shorttitle = {{{kPAM}} 2.0},
  author = {Gao, Wei and Tedrake, Russ},
  year = 2021,
  month = apr,
  journal = {IEEE Robotics and Automation Letters},
  volume = {6},
  number = {2},
  pages = {2962--2969},
  issn = {2377-3766, 2377-3774},
  doi = {10.1109/LRA.2021.3062315},
  copyright = {https://ieeexplore.ieee.org/Xplorehelp/downloads/license-information/IEEE.html}
}

@article{gaoKVILKeypointsBasedVisual2023,
  title = {K-{{VIL}}: {{Keypoints-Based Visual Imitation Learning}}},
  shorttitle = {K-{{VIL}}},
  author = {Gao, Jianfeng and Tao, Zhi and Jaquier, No{\'e}mie and Asfour, Tamim},
  year = 2023,
  month = oct,
  journal = {IEEE Transactions on Robotics},
  volume = {39},
  number = {5},
  pages = {3888--3908},
  issn = {1552-3098, 1941-0468},
  doi = {10.1109/TRO.2023.3286074}
}

@misc{houAdaptiveCompliancePolicy2025,
  title = {Adaptive {{Compliance Policy}}: {{Learning Approximate Compliance}} for {{Diffusion Guided Control}}},
  shorttitle = {Adaptive {{Compliance Policy}}},
  author = {Hou, Yifan and Liu, Zeyi and Chi, Cheng and Cousineau, Eric and Kuppuswamy, Naveen and Feng, Siyuan and Burchfiel, Benjamin and Song, Shuran},
  year = 2025,
  month = mar,
  number = {arXiv:2410.09309},
  eprint = {2410.09309},
  primaryclass = {cs},
  publisher = {arXiv},
  doi = {10.48550/arXiv.2410.09309},
  archiveprefix = {arXiv}
}

@misc{jaquierTransferLearningRobotics2023,
  title = {Transfer {{Learning}} in {{Robotics}}: {{An Upcoming Breakthrough}}? {{A Review}} of {{Promises}} and {{Challenges}}},
  shorttitle = {Transfer {{Learning}} in {{Robotics}}},
  author = {Jaquier, No{\'e}mie and Welle, Michael C. and Gams, Andrej and Yao, Kunpeng and Fichera, Bernardo and Billard, Aude and Ude, Ale{\v s} and Asfour, Tamim and Kragic, Danica},
  year = 2023,
  month = dec,
  number = {arXiv:2311.18044},
  eprint = {2311.18044},
  primaryclass = {cs},
  publisher = {arXiv},
  archiveprefix = {arXiv}
}

@misc{lakshmipathyContactEditArtist2023,
  title = {Contact {{Edit}}: {{Artist Tools}} for {{Intuitive Modeling}} of {{Hand-Object Interactions}}},
  shorttitle = {Contact {{Edit}}},
  author = {Lakshmipathy, Arjun S. and Feng, Nicole and Lee, Yu Xi and Mahler, Moshe and Pollard, Nancy S.},
  year = 2023,
  month = may,
  number = {arXiv:2305.02051},
  eprint = {2305.02051},
  primaryclass = {cs},
  publisher = {arXiv},
  doi = {10.48550/arXiv.2305.02051},
  archiveprefix = {arXiv}
}

@article{lipmanBiharmonicDistance2010,
  title = {Biharmonic Distance},
  author = {Lipman, Yaron and Rustamov, Raif M. and Funkhouser, Thomas A.},
  year = 2010,
  month = jun,
  journal = {ACM Transactions on Graphics},
  volume = {29},
  number = {3},
  pages = {1--11},
  issn = {0730-0301, 1557-7368},
  doi = {10.1145/1805964.1805971}
}

@misc{manuelliKPAMKeyPointAffordances2019,
  title = {{{kPAM}}: {{KeyPoint Affordances}} for {{Category-Level Robotic Manipulation}}},
  shorttitle = {{{kPAM}}},
  author = {Manuelli, Lucas and Gao, Wei and Florence, Peter and Tedrake, Russ},
  year = 2019,
  month = oct,
  number = {arXiv:1903.06684},
  eprint = {1903.06684},
  primaryclass = {cs},
  publisher = {arXiv},
  doi = {10.48550/arXiv.1903.06684},
  archiveprefix = {arXiv}
}

@article{markleyAveragingQuaternions2007,
  title = {Averaging {{Quaternions}}},
  author = {Markley, F. Landis and Cheng, Yang and Crassidis, John L. and Oshman, Yaakov},
  year = 2007,
  month = jul,
  journal = {Journal of Guidance, Control, and Dynamics},
  volume = {30},
  number = {4},
  pages = {1193--1197},
  issn = {0731-5090, 1533-3884},
  doi = {10.2514/1.28949}
}

@misc{millerDifferentialWalkSpheres2024,
  title = {Differential {{Walk}} on {{Spheres}}},
  author = {Miller, Bailey and Sawhney, Rohan and Crane, Keenan and Gkioulekas, Ioannis},
  year = 2024,
  month = may,
  number = {arXiv:2405.12964},
  eprint = {2405.12964},
  primaryclass = {cs},
  publisher = {arXiv},
  archiveprefix = {arXiv}
}

@article{muchachoAdaptiveDistanceFunctions,
  title = {Adaptive {{Distance Functions}} via {{Kelvin Transformation}} and {{Applications}} to {{Robotic Manipulation}}},
  author = {Muchacho, Rafael I Cabral and Pokorny, Florian T}
}

@misc{muchachoWalkSpheresPDEbased2024,
  title = {Walk on {{Spheres}} for {{PDE-based Path Planning}}},
  author = {Muchacho, Rafael I. Cabral and Pokorny, Florian T.},
  year = 2024,
  month = jun,
  number = {arXiv:2406.01713},
  eprint = {2406.01713},
  primaryclass = {cs},
  publisher = {arXiv},
  archiveprefix = {arXiv}
}

@article{oquabDINOv2LearningRobust,
  title = {{{DINOv2}}: {{Learning Robust Visual Features}} without {{Supervision}}},
  author = {Oquab, Maxime and Darcet, Timoth{\'e}e and Moutakanni, Th{\'e}o and Vo, Huy V and Szafraniec, Marc and Khalidov, Vasil and Fernandez, Pierre and Haziza, Daniel and Massa, Francisco and {El-Nouby}, Alaaeldin and Assran, Mahmoud and Ballas, Nicolas and Galuba, Wojciech and Howes, Russell and Huang, Po-Yao and Li, Shang-Wen and Misra, Ishan and Rabbat, Michael and Sharma, Vasu and Synnaeve, Gabriel and Xu, Hu and Jegou, Herv{\'e} and Mairal, Julien and Labatut, Patrick and Joulin, Armand and Bojanowski, Piotr}
}

@article{ovsjanikovFunctionalMapsFlexible2012,
  title = {Functional Maps: A Flexible Representation of Maps between Shapes},
  shorttitle = {Functional Maps},
  author = {Ovsjanikov, Maks and {Ben-Chen}, Mirela and Solomon, Justin and Butscher, Adrian and Guibas, Leonidas},
  year = 2012,
  month = aug,
  journal = {ACM Transactions on Graphics},
  volume = {31},
  number = {4},
  pages = {1--11},
  issn = {0730-0301, 1557-7368},
  doi = {10.1145/2185520.2185526}
}

@misc{renGroundedSAMAssembling2024,
  title = {Grounded {{SAM}}: {{Assembling Open-World Models}} for {{Diverse Visual Tasks}}},
  shorttitle = {Grounded {{SAM}}},
  author = {Ren, Tianhe and Liu, Shilong and Zeng, Ailing and Lin, Jing and Li, Kunchang and Cao, He and Chen, Jiayu and Huang, Xinyu and Chen, Yukang and Yan, Feng and Zeng, Zhaoyang and Zhang, Hao and Li, Feng and Yang, Jie and Li, Hongyang and Jiang, Qing and Zhang, Lei},
  year = 2024,
  month = jan,
  number = {arXiv:2401.14159},
  eprint = {2401.14159},
  primaryclass = {cs},
  publisher = {arXiv},
  doi = {10.48550/arXiv.2401.14159},
  archiveprefix = {arXiv}
}

@article{sawhneyMonteCarloGeometry,
  title = {Monte {{Carlo Geometry Processing}}:{{A Grid-Free Approach}} to {{PDE-Based Methods}} on {{Volumetric Domains}}},
  author = {Sawhney, Rohan and Crane, Keenan},
  volume = {38},
  number = {4}
}

@inproceedings{scherzingerForwardDynamicsCompliance2017,
  title = {Forward {{Dynamics Compliance Control}} ({{FDCC}}): {{A}} New Approach to Cartesian Compliance for Robotic Manipulators},
  shorttitle = {Forward {{Dynamics Compliance Control}} ({{FDCC}})},
  booktitle = {2017 {{IEEE}}/{{RSJ International Conference}} on {{Intelligent Robots}} and {{Systems}} ({{IROS}})},
  author = {Scherzinger, Stefan and Roennau, Arne and Dillmann, Rudiger},
  year = 2017,
  month = sep,
  pages = {4568--4575},
  publisher = {IEEE},
  address = {Vancouver, BC},
  doi = {10.1109/IROS.2017.8206325},
  isbn = {978-1-5386-2682-5}
}

@article{sharpDiffusionNetDiscretizationAgnostic2022,
  title = {{{DiffusionNet}}: {{Discretization Agnostic Learning}} on {{Surfaces}}},
  shorttitle = {{{DiffusionNet}}},
  author = {Sharp, Nicholas and Attaiki, Souhaib and Crane, Keenan and Ovsjanikov, Maks},
  year = 2022,
  month = mar,
  journal = {ACM Transactions on Graphics},
  volume = {41},
  number = {3},
  pages = {27:1--27:16},
  issn = {0730-0301},
  doi = {10.1145/3507905}
}

@article{sharpLaplacianNonmanifoldTriangle2020,
  title = {A {{Laplacian}} for {{Nonmanifold Triangle Meshes}}},
  author = {Sharp, Nicholas and Crane, Keenan},
  year = 2020,
  journal = {Computer Graphics Forum},
  volume = {39},
  number = {5},
  pages = {69--80},
  issn = {1467-8659},
  doi = {10.1111/cgf.14069}
}

@article{sharpVectorHeatMethod2019,
  title = {The {{Vector Heat Method}}},
  author = {Sharp, Nicholas and Soliman, Yousuf and Crane, Keenan},
  year = 2019,
  month = jun,
  journal = {ACM Transactions on Graphics},
  volume = {38},
  number = {3},
  pages = {1--19},
  issn = {0730-0301, 1557-7368},
  doi = {10.1145/3243651}
}

@misc{shawLEAPHandLowCost2023,
  title = {{{LEAP Hand}}: {{Low-Cost}}, {{Efficient}}, and {{Anthropomorphic Hand}} for {{Robot Learning}}},
  shorttitle = {{{LEAP Hand}}},
  author = {Shaw, Kenneth and Agarwal, Ananye and Pathak, Deepak},
  year = 2023,
  month = sep,
  number = {arXiv:2309.06440},
  eprint = {2309.06440},
  primaryclass = {cs},
  publisher = {arXiv},
  doi = {10.48550/arXiv.2309.06440},
  archiveprefix = {arXiv}
}

@inproceedings{simeonovNeuralDescriptorFields2022,
  title = {Neural {{Descriptor Fields}}: {{SE}}(3)-{{Equivariant Object Representations}} for {{Manipulation}}},
  shorttitle = {Neural {{Descriptor Fields}}},
  booktitle = {2022 {{International Conference}} on {{Robotics}} and {{Automation}} ({{ICRA}})},
  author = {Simeonov, Anthony and Du, Yilun and Tagliasacchi, Andrea and Tenenbaum, Joshua B. and Rodriguez, Alberto and Agrawal, Pulkit and Sitzmann, Vincent},
  year = 2022,
  month = may,
  pages = {6394--6400},
  publisher = {IEEE},
  address = {Philadelphia, PA, USA},
  doi = {10.1109/ICRA46639.2022.9812146},
  copyright = {https://doi.org/10.15223/policy-029},
  isbn = {978-1-7281-9681-7}
}

@misc{simeonovSE3EquivariantRelationalRearrangement2022,
  title = {{{SE}}(3)-{{Equivariant Relational Rearrangement}} with {{Neural Descriptor Fields}}},
  author = {Simeonov, Anthony and Du, Yilun and {Yen-Chen}, Lin and Rodriguez, Alberto and Kaelbling, Leslie Pack and {Lozano-Perez}, Tomas and Agrawal, Pulkit},
  year = 2022,
  month = nov,
  number = {arXiv:2211.09786},
  eprint = {2211.09786},
  primaryclass = {cs},
  publisher = {arXiv},
  doi = {10.48550/arXiv.2211.09786},
  archiveprefix = {arXiv}
}

@inproceedings{soCITRCoordinateInvariantTask2024,
  title = {{{CITR}}: {{A Coordinate-Invariant Task Representation}} for {{Robotic Manipulation}}},
  shorttitle = {{{CITR}}},
  booktitle = {2024 {{IEEE International Conference}} on {{Robotics}} and {{Automation}} ({{ICRA}})},
  author = {So, Peter and Cabral Muchacho, Rafael I. and Jeanne Kirschner, Robin and Swikir, Abdalla and Figueredo, Luis and {Abu-Dakka}, Fares J. and Haddadin, Sami},
  year = 2024,
  month = may,
  pages = {17501--17507},
  publisher = {IEEE},
  address = {Yokohama, Japan},
  doi = {10.1109/ICRA57147.2024.10611312},
  copyright = {https://doi.org/10.15223/policy-029},
  isbn = {979-8-3503-8457-4}
}

@article{tiGeometricOptimalControl2023,
  title = {A Geometric Optimal Control Approach for Imitation and Generalization of Manipulation Skills},
  author = {Ti, Boyang and Razmjoo, Amirreza and Gao, Yongsheng and Zhao, Jie and Calinon, Sylvain},
  year = 2023,
  month = jun,
  journal = {Robotics and Autonomous Systems},
  volume = {164},
  pages = {104413},
  issn = {09218890},
  doi = {10.1016/j.robot.2023.104413}
}

@article{urecheConstraintsExtractionAsymmetrical2018,
  title = {Constraints Extraction from Asymmetrical Bimanual Tasks and Their Use in Coordinated Behavior},
  author = {Ureche, Lucia Pais and Billard, Aude},
  year = 2018,
  month = may,
  journal = {Robotics and Autonomous Systems},
  volume = {103},
  pages = {222--235},
  issn = {09218890},
  doi = {10.1016/j.robot.2017.12.011}
}

@article{urecheTaskParameterizationUsing2015,
  title = {Task {{Parameterization Using Continuous Constraints Extracted From Human Demonstrations}}},
  author = {Ureche, Ana Lucia Pais and Umezawa, Keisuke and Nakamura, Yoshihiko and Billard, Aude},
  year = 2015,
  month = dec,
  journal = {IEEE Transactions on Robotics},
  volume = {31},
  number = {6},
  pages = {1458--1471},
  issn = {1552-3098, 1941-0468},
  doi = {10.1109/TRO.2015.2495003},
  copyright = {https://ieeexplore.ieee.org/Xplorehelp/downloads/license-information/IEEE.html}
}

@misc{vedoveMeshDMPMotionPlanning2024,
  title = {{{MeshDMP}}: {{Motion Planning}} on {{Discrete Manifolds}} Using {{Dynamic Movement Primitives}}},
  shorttitle = {{{MeshDMP}}},
  author = {Vedove, Matteo Dalle and {Abu-Dakka}, Fares J. and Palopoli, Luigi and Fontanelli, Daniele and Saveriano, Matteo},
  year = 2024,
  month = oct,
  number = {arXiv:2410.15123},
  eprint = {2410.15123},
  primaryclass = {cs},
  publisher = {arXiv},
  doi = {10.48550/arXiv.2410.15123},
  archiveprefix = {arXiv}
}

@article{vochtenGeneralizingDemonstratedMotion2019,
  title = {Generalizing Demonstrated Motion Trajectories Using Coordinate-Free Shape Descriptors},
  author = {Vochten, Maxim and De Laet, Tinne and De Schutter, Joris},
  year = 2019,
  month = dec,
  journal = {Robotics and Autonomous Systems},
  volume = {122},
  pages = {103291},
  issn = {09218890},
  doi = {10.1016/j.robot.2019.103291}
}

@article{vochtenInvariantDescriptorsMotion2023,
  title = {Invariant {{Descriptors}} of {{Motion}} and {{Force Trajectories}} for {{Interpreting Object Manipulation Tasks}} in {{Contact}}},
  author = {Vochten, Maxim and Mohammadi, Ali Mousavi and Verduyn, Arno and De Laet, Tinne and Aertbeli{\"e}n, Erwin and De Schutter, Joris},
  year = 2023,
  month = dec,
  journal = {IEEE Transactions on Robotics},
  volume = {39},
  number = {6},
  pages = {4892--4912},
  issn = {1552-3098, 1941-0468},
  doi = {10.1109/TRO.2023.3309230},
  copyright = {https://creativecommons.org/licenses/by/4.0/legalcode}
}

@misc{wangD$^3$FieldsDynamic3D2024a,
  title = {D\$\textasciicircum 3\${{Fields}}: {{Dynamic 3D Descriptor Fields}} for {{Zero-Shot Generalizable Rearrangement}}},
  shorttitle = {D\$\textasciicircum 3\${{Fields}}},
  author = {Wang, Yixuan and Zhang, Mingtong and Li, Zhuoran and Kelestemur, Tarik and {Driggs-Campbell}, Katherine and Wu, Jiajun and {Fei-Fei}, Li and Li, Yunzhu},
  year = 2024,
  month = oct,
  number = {arXiv:2309.16118},
  eprint = {2309.16118},
  primaryclass = {cs},
  publisher = {arXiv},
  doi = {10.48550/arXiv.2309.16118},
  archiveprefix = {arXiv}
}

@article{xueReactiveDiffusionPolicy,
  title = {Reactive {{Diffusion Policy}}:},
  author = {Xue, Han and Ren, Jieji and Chen, Wendi and Zhang, Gu and Fang, Yuan and Gu, Guoying and Xu, Huazhe and Lu, Cewu}
}

@misc{ze3DDiffusionPolicy2024,
  title = {{{3D Diffusion Policy}}: {{Generalizable Visuomotor Policy Learning}} via {{Simple 3D Representations}}},
  shorttitle = {{{3D Diffusion Policy}}},
  author = {Ze, Yanjie and Zhang, Gu and Zhang, Kangning and Hu, Chenyuan and Wang, Muhan and Xu, Huazhe},
  year = 2024,
  month = sep,
  number = {arXiv:2403.03954},
  eprint = {2403.03954},
  primaryclass = {cs},
  publisher = {arXiv},
  doi = {10.48550/arXiv.2403.03954},
  archiveprefix = {arXiv}
}

@misc{schulman2017proximalpolicyoptimizationalgorithms,
      title={Proximal Policy Optimization Algorithms}, 
      author={John Schulman and Filip Wolski and Prafulla Dhariwal and Alec Radford and Oleg Klimov},
      year={2017},
      eprint={1707.06347},
      archivePrefix={arXiv},
      primaryClass={cs.LG},
      url={https://arxiv.org/abs/1707.06347}, 
}
\bibliographystyle{sciencemag}

%
%
%
%
%
%


\section*{Acknowledgments}
We thank L. Beber for providing the soft object samples used in our experiments and for his assistance with robot integration and camera calibration. We also thank J. Maceiras for his support with the force/torque sensor integration and camera calibration. We acknowledge A. Razmjoo, J. Hermus, M. Schoenger, A. Maric, Y. Li, and Y. Zhang for their valuable discussions. We used ChatGPT to assist in revising and refining the language of this manuscript. We used Claude Code to assist in organizing and documenting the software repositories.
\paragraph*{Funding:}
This work was supported by the State Secretariat for Education, Research and Innovation in Switzerland for participation in the European Commission's Horizon Europe Program through the INTELLIMAN project (\url{https://intelliman-project.eu/}, HORIZON-CL4-Digital-Emerging Grant 101070136) and the SESTOSENSO project (\url{http://sestosenso.eu/}, HORIZON-CL4-Digital-Emerging Grant 101070310).
\paragraph*{Author contributions:}
Conceptualization,  simulation, experiments, analysis, visualization, and writing: C.B. discussions, supervision, robot integration, and revision: T.L. initial task scope, resources, supervision, and revision: S.C. 
\paragraph*{Competing interests:}
There are no competing interests to declare.
\paragraph*{Data and materials availability:}
the data for this study will be deposited at: 

\url{https://github.com/idiap/diffused_fields}


\subsection*{Supplementary materials}
Materials and Methods\\
Supplementary Results\\
Figures S1 to S7\\
Tables S1 to S3


\newpage


\renewcommand{\thefigure}{S\arabic{figure}}
\renewcommand{\thetable}{S\arabic{table}}
\renewcommand{\theequation}{S\arabic{equation}}
\renewcommand{\thepage}{S\arabic{page}}
\setcounter{figure}{0}
\setcounter{table}{0}
\setcounter{equation}{0}
\setcounter{page}{1} 


\begin{center}
\section{Supplementary Materials for\\ \scititle}


	Cem Bilaloglu$^{1,2\ast}$,
	Tobias Löw$^{1,2}$,
	Sylvain Calinon$^{1,2}$\\
	\small$^{1}$Idiap Research Institute, 1920 Martigny, Switzerland, 
	\small$^{2}$EPFL, 1015 Lausanne, Switzerland\\
	\small$^\ast$Corresponding author. Email: cem.bilaloglu@epfl.ch

\end{center}

\subsubsection{This PDF file includes:}
Materials and Methods\\
Supplementary Results\\
Figures S1 to S7\\
Tables S1 to S3\\

\subsubsection{Other Supplementary Materials for this manuscript:}

\newpage

\section{Materials and Methods}

\subsection{Keypoint Extraction}
Keypoints are inputs to our representation and can be obtained by automatic extraction, transfer with a learning based approach, or manual annotation. As shown in Fig. 6E, our method is robust to noise in keypoint extraction. In real-world experiments  we used automatic keypoint extraction. As a proof-of-concept we also performed keypoint transfer using pretrained vision encoders~\cite{oquabDINOv2LearningRobust}.

\subsubsection{Automatic Keypoint Extraction}
For automatic keypoint extraction we assume the observed point cloud has an open boundary, which is the typical case for single camera acquisition. 

For slicing and peeling experiments, the keypoints are the antipodal poles of the object, namely the stem end and the tip.
We detect the two ends of elongated fruits and vegetables such as banana, pear, or cucumber with an iterative furthest point strategy on the point cloud driven by the scalar diffusion equation~\eqref{eq:diffusion}. If the point cloud is not elongated, the method may be unreliable and we recommend manual annotation.

First, we compute the geometric center of the surface and use it as a source similar to~\eqref{eq:sources_sinks}. Next, we solve~\eqref{eq:implicit_time_stepping} to diffuse a scalar field over the surface. The vertex with the minimum diffused value is selected as the first end point since it lies far along the intrinsic geodesic from the center. Then, we use the first detected end point as the new source and repeat the process and select the minimum value vertex as the second end point.

On a set of fifty randomly deformed banana point clouds with combined scaling, twisting, and bending, the two stage diffusion consistently identified anatomically correct ends at the stem and the tip (Fig.~\ref{fig:keypoint_extraction}A, i). The method requires no training data and transfers to other elongated objects since it relies only on intrinsic geometry.

In coverage experiments, we set the object boundary as the keypoints. Accordingly, one can use any boundary estimation procedure. We use a procedure inspired by the Point Cloud Library's \emph{boundaryEstimation} function. We tested this approach with point clouds that we collected from various objects (Fig.~\ref{fig:keypoint_extraction}A, ii).
\subsubsection{Keypoint Transfer}
We used a pretrained vision encoder DINOv2~\cite{oquabDINOv2LearningRobust} to transfer keypoints annotated on one object to a target object encountered at runtime. This method requires no training, fine-tuning, or dataset-specific optimization. It relies entirely on the general-purpose visual understanding encoded in pretrained foundation models. 

\begin{figure}[]
\centering
\includegraphics[width=1.0\linewidth]{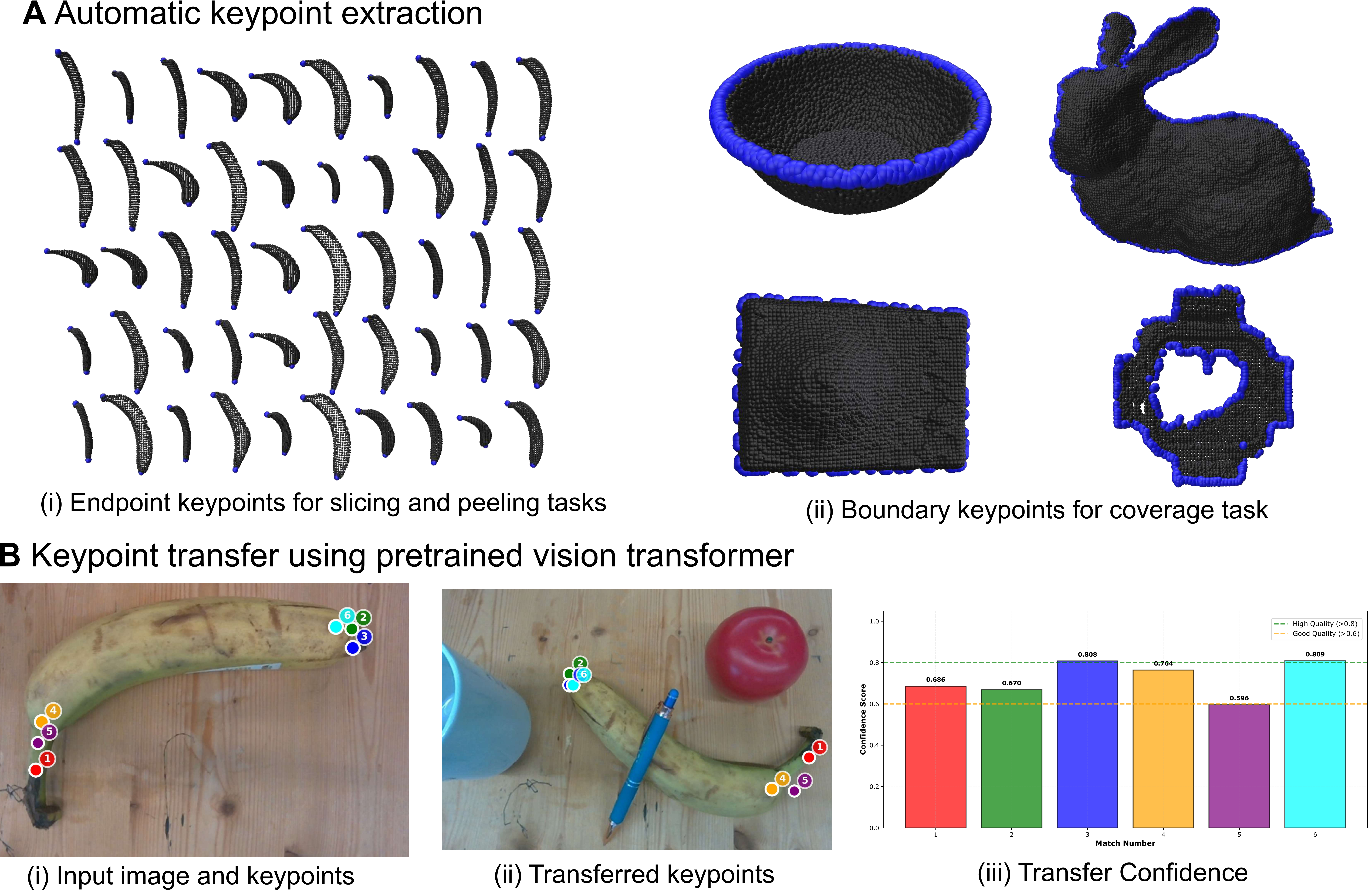}
\caption{\textbf{Keypoint extraction}. (\textbf{A}) Automatic keypoint extraction for slicing, peeling and coverage experiments. Point cloud is in black and keypoints are in blue. (i) 50 twisted, bent and anisotropically scaled instances of banana point cloud used for testing our automatic keypoint extraction routine. For slicing and peeling tasks, keypoints need to be at two ends of the object. (ii) Various real-world point clouds tested with our boundary keypoint extraction routine. For coverage tasks, the keypoints need to span the object boundary. (\textbf{B}) Keypoint transfer using pretrained vision transformer~\cite{oquabDINOv2LearningRobust}. (i) User annotated keypoints on the input image. (ii) Zero-shot transfer of annotated keypoints to the image collected at runtime. (iii) Transfer confidence for each keypoint.}
\label{fig:keypoint_extraction}
\end{figure}

The core principle is that semantically similar image regions should produce similar feature representations, even under substantial appearance variations. For each query keypoint in the source image, we extract a local feature descriptor by processing a surrounding image patch through the DINOv2 encoder. We then search the target image for the location that maximizes feature similarity, measured via cosine similarity in the learned feature space. We provided the results of our proof-of-concept experiment in Fig.~\ref{fig:keypoint_extraction}B.

\subsection{Local Action Primitives}
Local action primitives are simple closed-loop policies expressed in object-centric local reference frames represented by DOF. At each trajectory point $\bm{x}_t$, the controller queries the WoS to obtain a local orthonormal coordinate frame  $\mathbf{F}_{\bm{x}} = [\mathbf{u}, \mathbf{v}, \mathbf{n}] \in SO(3)$ from DOF, where
$\mathbf{u}$ is the direction towards the keypoints parallel to the surface (gradient of the smoothened geodesic distance to the keypoints along the surface levelsets),
$\mathbf{v}$ is the direction orthogonal to $\mathbf{u}$ and $\mathbf{v}$ (preserves the smoothened geodesic distance to the keypoints on the surface levelsets and smoothened Euclidean distance to the surface),
and $\mathbf{n}$ is the direction towards the surface (gradient of the smoothened distance field to the surface).

The controller computes the next position by stepping according to the action $\bm{a}_t$ expressed in the local reference frame 
\begin{equation}
	\bm{x}_{t+1} = \bm{x}_t + \delta\mathbf{F}_{\bm{x}}\bm{a}_t,
\end{equation}
where $\delta$ is the step size. Each primitive selects the action based on its motion phase (slide along, go down, lift up, and so on), as shown in Fig. 3B. The transition between phases is determined by the distance to the keypoints and by the distance to the surface.

\subsection{Local Policy Learning}

Local action primitives can also be learned as local policies with reinforcement learning. In our experiments, we represented the policy using a multi-layer perceptron that was trained using proximal policy optimization (PPO) \cite{schulman2017proximalpolicyoptimizationalgorithms} from \textit{stable\_baselines3}.\footnote{https://github.com/DLR-RM/stable-baselines3} During training we used the following dense reward function
\begin{equation}\label{eq:reward_function}
	\begin{split}
		r_t(\bm{s}_t,\bm{a}_t) = & f_1(\bm{s}_t,\bm{a}_t) + f_2(\bm{s}_t,\bm{a}_t) + f_3(\bm{s}_t,\bm{a}_t) + f_4(\bm{s}_t,\bm{a}_t) + f_5(\bm{s}_t,\bm{a}_t)
        \\
        & f_1(\bm{s}_t,\bm{a}_t) = -0.1 \quad \text{time penalty (accumulating)}
        \\
        & f_2(\bm{s}_t,\bm{a}_t) =  -0.01 \|\bm{a}\|^2 \quad \text{action penalty}
        \\
        & f_3(\bm{s}_t,\bm{a}_t) = \begin{cases}
            10 (d_c(\bm{s}_{t-1}) - d_c(\bm{s}_{t})) \quad \text{off circle}
            \\
            0.5 + 10 (d_t(\bm{s}_{t-1}) - d_t(\bm{s}_{t})) \quad \text{on circle}
        \end{cases}
        \\
        & f_4(\bm{s}_t,\bm{a}_t) = -5 d_c(\bm{s}_{t}) \quad \text{penetration penalty}      
	\end{split},
\end{equation}
where $s_t$ and $a_t$ are the states and actions at time $t$, $d_c(\cdot)$ is the distance to the circle and $d_t(\cdot)$ is the distance to the target. When the target is reached, we additionally give a single sparse reward
\begin{equation}
    r_f(\bm{s}_t) = 10 \quad \text{target reached reward (one time)}.
\end{equation}
To speed up training, we pretrain with 10 demonstrations, then train the policy for $5\times 10^5$ timesteps with the entropy coefficient annealed linearly from 0.04 to 0.002, which encourages initial exploration and convergence to a near deterministic policy. For the local policy, we compute DOF using a keypoint at the target position.

\subsection{Diffused Orientation Fields from Directed and Oriented Keypoints}
Complementary to the scalar diffusion method described in the main text for computing diffused orientation field, we provide two alternatives that utilize directed and oriented keypoints. 




\subsubsection{Orientation Fields via Diffused Directed Keypoints}
\label{subsec:directed_keypoints}
Solving scalar diffusion and computing its gradient yields directions that always points towards or away from the keypoints. However, in some applications, we might directly have access to direction information at keypoints that do not necessarily correspond to sources or sinks. For example, in a wiping task, we might want to specify the desired wiping direction at specific locations on the object surface and smoothly interpolate these directions.

Let \(\Omega \subset \mathbb{R}^n\) and a vector field \(\mathbf{v}(\boldsymbol{x},\tau) = (v_1,v_2, \ldots v_m)\). One can compute a diffused vector field by solving 
$m$ independent scalar diffusion problems 
\begin{equation}
\dot{v}_i = \Delta v_i \quad \text{in } \Omega,
\qquad
v_i\big|_{\partial\Omega} = g_i(\boldsymbol{x}),
\qquad i=\{1,2,\ldots, m\}.
\label{eq:vector_diffusion}
\end{equation}
where $g_i(\boldsymbol{x})$ is the $i^{th}$ component of the directed keypoints. Since diffusion would result in magnitude decay and we are interested in direction vectors, we normalize the diffused vector field to obtain the direction field.

For vectors on curved surfaces, an additional constraint arises: vectors must reside in the tangent bundle, $\mathbf{v} \in \mathcal{T} \mathcal{M}$, which is the manifold formed by the union of all tangent spaces of the original manifold $\mathcal{M}$. Consequently, the components of a vector field cannot be diffused independently, as curved manifolds lack a global coordinate system, and vectors at different points cannot be directly compared. To address this, one can use the connection Laplacian $\Delta_{\mathcal{M}}^{\nabla}$, defined as the composition of the covariant derivative and its adjoint for diffusing a tangent vector field $\mathbf{v}(\boldsymbol{x}, \tau)$:

\begin{equation}
\dot{\mathbf{v}}(\boldsymbol{x}, \tau)=\Delta_{\mathcal{M}}^{\nabla} \mathbf{v}(\boldsymbol{x}, \tau) .
\end{equation}


Similar to the Euclidean setting, one needs to normalize the diffused vector field to obtain the direction field on the manifold. We refer the reader to~\cite{sharpVectorHeatMethod2019} for more details on vector diffusion for curved surfaces.

\subsubsection{Orientation Fields via Diffused Oriented Keypoints}
\label{subsec:oriented_keypoints}
Scalar and vector diffusion yield local reference frames where the z-axis aligns with the surface normal, and the x–y plane is tangent to the surface. While this is often useful, some applications might require diffusing desired full 3D orientations, i.e.~oriented keypoints, rather than surface-constrained directions. For example, one can specify the full desired tool orientation at specific locations on the object and use orientation diffusion to smoothly interpolate these orientations over the entire workspace.

Notably, in order to diffuse full 3D orientations over a surface, we cannot use the method that we presented in the main paper for workspace diffusion as the Walk on Spheres method is limited to Euclidean spaces and cannot be used on curved surfaces. Also we cannot use the scalar or vector diffusion methods presented in the previous sections since they cannot represent full 3D orientations and $\mathrm{SO}(3)$ is not a vector space.
To diffuse full 3D orientations over a surface, we must choose a suitable representation. If we use rotation matrices to represent orientations, component-wise diffusion can violate the orthogonality and determinant constraints of $\mathrm{SO}(3)$. A better alternative is to diffuse columns of the rotation matrix which correspond to the axes of the local reference frames independently, as we discuss in more detail in the next section. However, this is still an approximation as we ignore the structure of $\mathrm{SO}(3)$ during the diffusion then correct it afterwards using re-orthonormalization. 

When having a few oriented keypoints, a more principled approach is to utilize the Lie algebra $\mathfrak{spin}(3)$, which forms a vector space over $\mathbb{R}$ and can be represented using pure quaternions (quaternions with zero scalar part). The associated Lie group $\mathrm{Spin}(3)$, represented by unit quaternions, is a double cover of $\mathrm{SO}(3)$, meaning that $\mathbf{q}$ and $-\mathbf{q}$ correspond to the same rotation. The exponential and logarithmic maps between $\mathfrak{spin}(3)$ and $\mathrm{Spin}(3)$ are given as:
\begin{equation}
	\exp{(\mathbf{u}\phi)} = \mathbf{q} = \cos(\phi) + \sin(\phi)\mathbf{u}, \label{eq:expmap}
\end{equation}
\begin{equation}
	\log{(\mathbf{q})} = \mathbf{u}\phi = \arccos(w) \frac{\mathbf{u}}{|\mathbf{u}|}. \label{eq:logmap}
\end{equation}

These maps allow computations to be performed in a linear vector space, enabling independent diffusion of each component, as we described in the previous section. However, similar to the vector diffusion case, one needs to compensate for the magnitude decay due to diffusion. Notably, we cannot just normalize the diffused pure quaternions since their magnitude is related to the rotation angle. As proposed in~\cite{sharpVectorHeatMethod2019}, the magnitude decay due to diffusion can be measured using an indicator function on the boundaries
\begin{equation}
\phi(\boldsymbol{x}, 0)=\left\{\begin{array}{ll}
1, & \boldsymbol{x} \in \partial \Omega \\
0, & \boldsymbol{x} \in \Omega
\end{array} ,\right.
\label{eq:indicator_function}
\end{equation}
and by diffusing the magnitude as a scalar field 
\begin{equation}
u(\boldsymbol{x}, 0)=\left\{\begin{array}{ll}
\|\mathbf{v}(\boldsymbol{x},\tau)\|, & \boldsymbol{x} \in \partial \Omega \\
0, & \boldsymbol{x} \in \Omega
\end{array} .\right.
\label{eq:magnitude_diffusion}
\end{equation}

Then, one can compensate for the magnitude decay of the diffused vector field $\mathbf{v}(\boldsymbol{x},t)$ 
\begin{equation}
\mathbf{\bar{v}}(\boldsymbol{x}, \tau)=\frac{\mathbf{v}(\boldsymbol{x}, \tau)u(\boldsymbol{x},\tau)}{\|\mathbf{v}(\boldsymbol{x}, \tau)\|\phi(\boldsymbol{x}, \tau)}.
\label{eq:magnitude_compensation}
\end{equation}

After the diffusion, we map back the diffused pure quaternions to unit quaternions. Our algorithm for diffusing orientations proceeds as follows:
\begin{enumerate}
	\item Represent orientations as unit quaternions $\mathbf{q}$.
    \item Ensure that all quaternions are in the same hemisphere by flipping their signs if necessary.
	\item Map each quaternion to a pure quaternion $\mathbf{u} \in \mathfrak{spin}(3)$ using the logarithmic map~\eqref{eq:logmap}.
	\item Impose these pure quaternions as Dirichlet boundary conditions in the diffusion equation~\eqref{eq:diffusion}.
	\item Solve the diffusion equation independently for each component of the pure quaternions following the procedure in \eqref{eq:vector_diffusion}.
	\item Diffuse the indicator function~\eqref{eq:indicator_function} and the magnitude function~\eqref{eq:magnitude_diffusion} and compensate for the magnitude decay using \eqref{eq:magnitude_compensation}.
	\item Map the diffused results back to unit quaternions using the exponential map~\eqref{eq:expmap}.
\end{enumerate}
Note that, we recommend this method only for diffusing a few oriented keypoints on the surface and to use the method presented in the main paper for workspace diffusion. 

\subsubsection{Approximated Diffused Orientation Field using Vector Diffusion and Orthonormalization}
In this section, we describe a simple approximation for computing diffused orientation fields on the workspace. We diffuse orthonormal direction vectors independently followed by orthonormalization. Let an orientation at a point be represented by a rotation matrix
\begin{equation}
\mathbf{R}=\big[\,\mathbf{r}_{x}\;\; \mathbf{r}_{y}\;\; \mathbf{r}_{z}\,\big]\in \mathrm{SO}(3),
\end{equation}
whose columns \(\mathbf{r}_{x},\mathbf{r}_{y},\mathbf{r}_{z}\in\mathbb{R}^{3}\) encode the local \(x,y,z\) axes. In order to diffuse individual column vectors, we independently solve three scalar diffusion equations as presented in~\eqref{eq:vector_diffusion} and normalize to unit magnitude. We repeat this process for all three column vectors and combine the results into a field of $3\times3$ matrices $\mathbf{M}(\boldsymbol{x})$.

\begin{figure}[]
\centering
\includegraphics[width=1.0\linewidth]{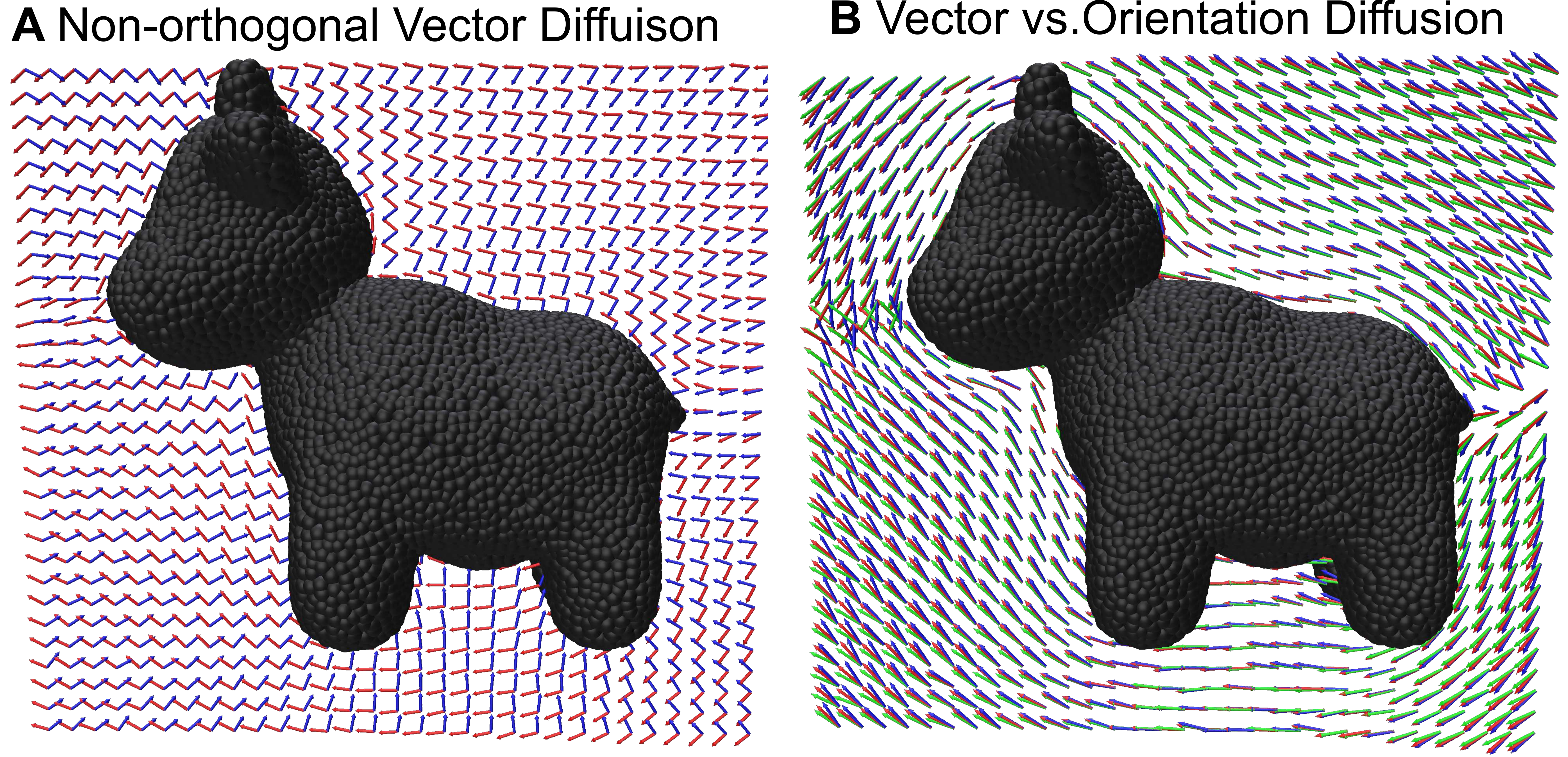}
\caption{\textbf{Diffusing the vectors composing the local frames independently}. (\textbf{A}) Diffusing vectors independently breaks orthogonality of orientations. (\textbf{B}) Comparison of the local reference frames computed by orthonormalized vector diffusion and orientation diffusion (our method). We visualize the x-axis of orientation diffusion in green, the x-axis of the orthonormalized vector diffusion using SVD orthogonalization in red and orthogonalization with Z-axis fixed in blue.}
\label{fig:vector_diffusion}
\end{figure}

Because these three problems are solved independently in a vector space, the result \(\mathbf{M}(\boldsymbol{x})\) generally violates the constraints of \(\mathrm{SO}(3)\):
\begin{equation}
\mathbf{M}(\boldsymbol{x})^{\!\top}\mathbf{M}(\boldsymbol{x})\neq \mathbf{I}
\quad\text{and}\quad
\det \mathbf{M}(\boldsymbol{x})\neq 1.
\end{equation}
We illustrate violations of orthogonality in Fig.~\ref{fig:vector_diffusion}A.

To convert the diffused matrix \(\mathbf{M}(\boldsymbol{x})\) into a valid orientation we orthonormalize it by projecting onto \(\mathrm{SO}(3)\). The projection that makes the smallest Frobenius norm change,
\begin{equation}
\mathbf{R}^{\star}(\mathbf{x})\;=\;\arg\min_{\mathbf{R}\in \mathrm{SO}(3)} \,\|\mathbf{M}(\mathbf{x})-\mathbf{R}\|_{F},
\end{equation}
is given by the polar decomposition computed via a thin singular value decomposition (SVD) \(\mathbf{M}=\mathbf{U}\boldsymbol{\Sigma}\mathbf{V}^{\!\top}\). This symmetric orthogonalization treats all columns of \(\mathbf{M}\) in a balanced way and yields an orthonormal frame with determinant one. A sequential alternative is the Gram-Schmidt process applied to the columns of \(\mathbf{M}\). However, the latter approach is order dependent and would introduce bias toward the first column and distort the other columns. We compare orthonormalized vector diffusion using both SVD and Gram-Schmidt with fixed z-direction to the orientation diffusion (Fig.~\ref{fig:vector_diffusion}B). For a quantitative comparison, we measured the smoothness of the resulting orientation fields using angular deviation between neighboring local x-directions. For each point on the grid visualized in Fig.~\ref{fig:vector_diffusion}, we computed the angle of neighboring local x-directions in the xz-plane with respect to a common reference direction. We present the statistics of the angular deviation in Table~\ref{tab:angular_deviation_stats}. As expected, we obtain the smoothest field using the orientation diffusion method presented in the main paper, since other methods do not directly consider diffusing orientations but they first consider smoothness of the vector diffusion and only later enforce orientation constraint. Although the SVD orthonormalized vector diffusion approach provides a very good approximation for the Spot point cloud, the difference might be higher for more complex shapes.
\begin{table}
	\centering
	\caption{\textbf{Orientation smoothness metrics across methods.}
	We report the average angular deviation, standard deviation of the angular deviation, and maximum angular deviation between neighboring local \(x\) directions. Note that the exact values can change slightly ($<1\%$) due to the randomness of the Walk on Spheres.}
	\label{tab:angular_deviation_stats}
	\begin{tabular}{lccc}
	\hline
	\textbf{Method} & \textbf{Average} & \textbf{Std. Dev.} & \textbf{Max. } \\
	\hline
	Orthonormalized Vector Diffusion (Z-fixed)       & 9.80$^\circ$  & 21.34$^\circ$ & 179.90$^\circ$ \\
	Orthonormalized Vector Diffusion (SVD)           & 7.12$^\circ$  & 12.11$^\circ$ & 160.19$^\circ$ \\
	Orientation Diffusion (Ours)  & 6.25$^\circ$  & 10.41$^\circ$ & 152.58$^\circ$ \\
	\hline
	\end{tabular}
\end{table}
\section{Supplementary Results}

\subsection{Computational Performance}
We evaluated the computational performance of our method on point clouds with varying numbers of vertices, using a laptop equipped with an Apple M1 Pro chip and 32 GB of RAM. The results are presented in Table~\ref{tab:performance}. 

As shown in Table~\ref{tab:performance}, preprocessing time is primarily dominated by the computation of the discrete Laplacian. Even in complex scenes (for example, those involving four distinct objects with 12k vertices each), the total preprocessing time remains under one second. Crucially, the discrete Laplacian is invariant under isometric transformations (e.g., rigid body motions or bending without stretching), which account for most of the transformations encountered in robotic manipulation. This makes DOF particularly well-suited to such settings, as the Laplacian does not need to be recomputed when the geometry undergoes these typical changes.

At runtime, DOF's value representing the local reference frame at the robot's position is computed using the Walk on Spheres (WoS) method. WoS is a Monte Carlo technique similar to ray casting, well-suited for massive parallelization and hardware acceleration. Despite being implemented in pure Python on the CPU, without adaptive step sizing, importance sampling, or boundary value caching, still our method enables real-time operation. Importantly, since WoS is inherently parallelizable, the runtime would remain constant with respect to the number of query points when executed on a GPU, making DOF evaluation highly scalable.

\begin{table}[]
    \centering
    \begin{tabular}{lccccc}
        \hline
        \textbf{Object} & \textbf{Points} & \textbf{Laplacian} & \textbf{Factorization} & \textbf{Surface Diffusion}  & \textbf{WoS} \\
        \hline
        Bowl   &  3,000  & 58 ms  & 10 ms  & 0.3 ms  & 4 ms  \\
        Spot   &  6,000  & 133 ms & 27 ms  & 0.6 ms  & 7 ms  \\
        Bunny  &  12,000 & 244 ms & 44 ms  & 1.1 ms  & 26 ms \\
        \hline
    \end{tabular}
    \caption{\textbf{Average computation times for the main components of our method.} For the Walk on Spheres (WoS), we used 256 samples per query and considered a sample to have converged when it was within twice the average neighbor distance from the boundary.}
    \label{tab:performance}
\end{table}

\subsection{Comparison with Alternative Baselines}
We compared our method to three alternative baselines for computing local reference frames in terms of smoothness. We selected these baselines to isolate different components of our method, allowing us to evaluate the importance of each component. We provide a detailed description of each baseline and the results of the comparisons below. 

\subsubsection{Nearest Frame Baseline}
\begin{figure}[]
\centering
\includegraphics[width=0.6\linewidth]{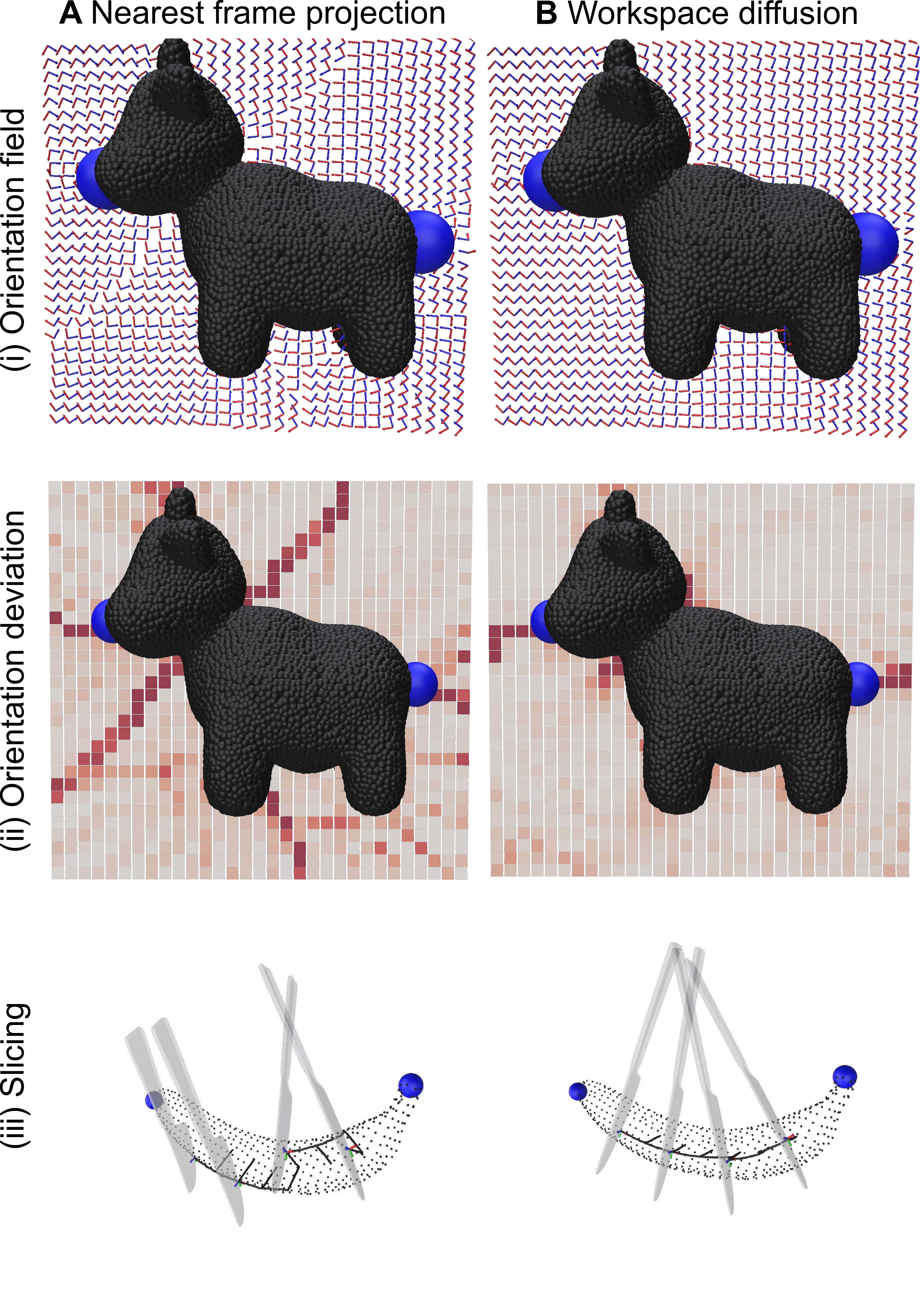}
\caption{\textbf{Comparison between nearest frame baseline and workspace diffusion}. (\textbf{A}) Nearest frame projection results in discontinuity in the orientation field since the nearby workspace points are mapped to distant points on the point cloud. (\textbf{B}) Workspace diffusion (ours) resulting in smooth orientation field. (i) A slice of the orientation field on Spot's symmetry plane. (ii) Deviation of the orientation in the local neighborhood shown using a heat map. (iii) Slicing task using nearest frame baseline resulting in undesired discontinuity due to the non-smooth orientation field.}
\label{fig:nearest_frame_baseline}
\end{figure}

Nearest frame baseline keeps the diffused surface orientation field but replaces the workspace diffusion with the nearest frame projection. Specifically, for each point in the workspace, we find the nearest point on the surface and use the local frame at that point as the local frame at the query point. We then compared our method (workspace diffusion) to the nearest frame baseline. We quantified smoothness using the average angular difference between neighboring local \(x\) directions. We measured $\approx 1.5$x higher average angular difference for the nearest frame baseline compared to our method, indicating that workspace diffusion produces a significantly smoother field. Note that this is an expected result since we can think of workspace diffusion as a smoothed projection to the boundary. We provided the results of this comparison in Fig.~\ref{fig:nearest_frame_baseline}. As we show Fig.~\ref{fig:nearest_frame_baseline}, iii, the nearest frame baseline produces discontinuities in the local frames, resulting in undesired behavior for the slicing task. Also, in other downstream robotics tasks, the discontinuities in the local frames would lead to force spikes and jerky motion.
    
\subsubsection{Vector Projection Baseline}
Vector projection baseline complements the nearest frame baseline as it provides a comparison for surface diffusion component of our approach. For computing the x-direction at a query point, we considered the weighted sum of the vectors connecting the query point to the keypoints. Next, we projected this vector to the tangent plane at the query point. One can compute the tangent plane at any point in the workspace, which is given by the gradient of the signed distance field to the surface.

\begin{figure}[]
\centering
\includegraphics[width=0.6\linewidth]{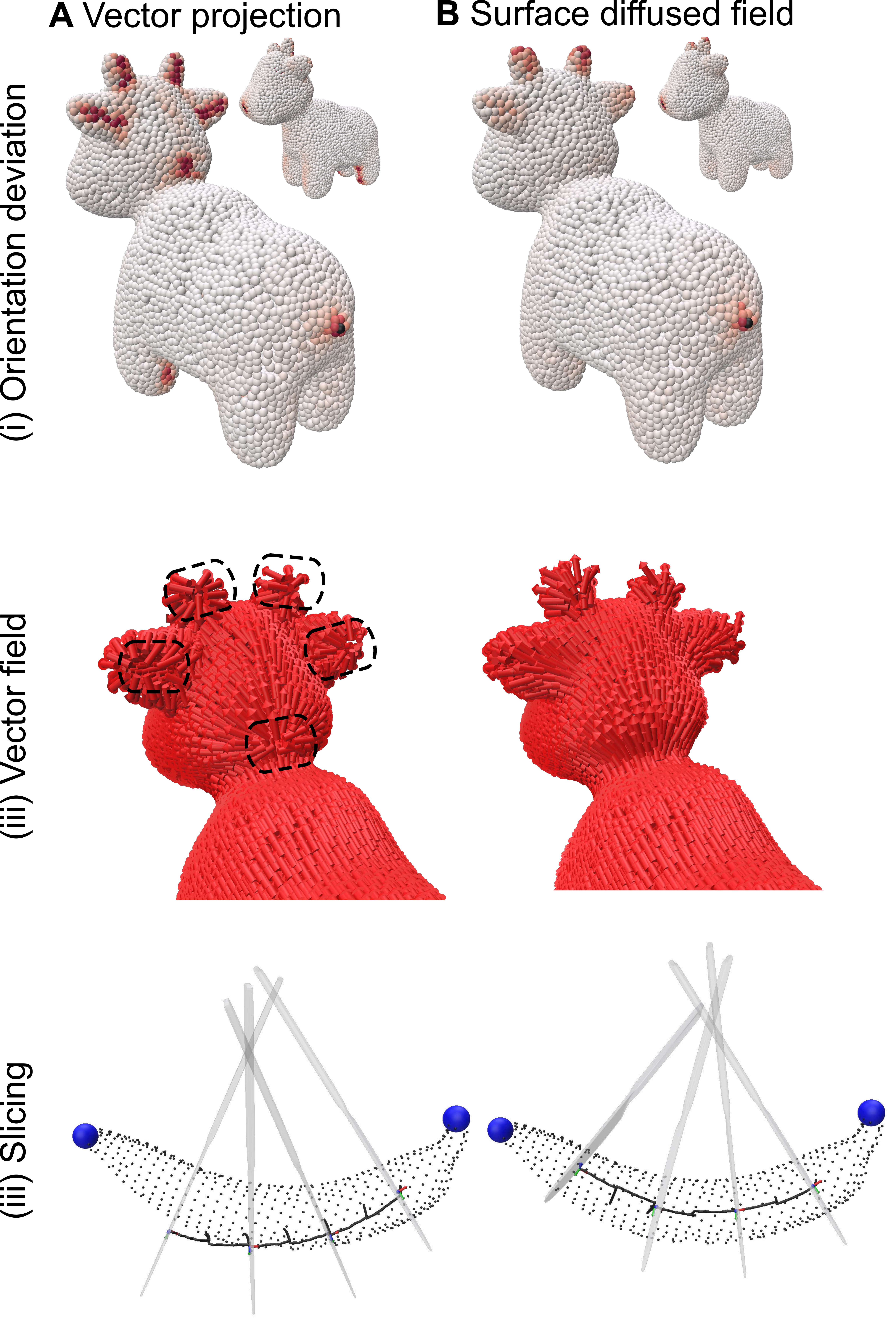}

\caption{\textbf{Comparison between vector projection baseline and surface diffusion}. (\textbf{A}) Vector projection baseline. Vector projection results in a non-smooth vector field on the tangent planes leading to spurious sources and sinks. (\textbf{B}) Surface diffusion (ours) producing a smooth vector field on the tangent planes free of sources and sinks except the ones at keypoints. (i) Orientation deviation on the tangent plane shown using a heat map. (ii) Deviation of the orientation in the local neighborhood. Higher red intensity show higher deviation. (iii) Slicing task comparison. We visualize the slicing trajectory with the black curve and visualize the tool pose in 4 equally-spaced time intervals.}
\label{fig:tangent_projection_baseline}
\end{figure}

We compared this baseline to our approach by using a number of spurious sources/sinks (i.e. undesired singularities) and visualized the results in Fig.~\ref{fig:tangent_projection_baseline}. We marked the spurious sources/sinks by using average angular difference between neighboring local \(x\) directions. As can be seen in Fig.~\ref{fig:tangent_projection_baseline}, i-ii, the vector projection baseline produced spurious sources/sinks (9 in total for the Spot point cloud) in the vector field, whereas our method is free of unaccounted singularities other than the ones resulting from keypoints. Note that, although it seems in Fig.~\ref{fig:tangent_projection_baseline}B, ii that our method produces a singularity in the horns of the Spot, this is an artifact of our angular deviation computation (error introduced by projecting neighboring vectors to the tangent space). We show this by visualizing the vector field in Fig.~\ref{fig:tangent_projection_baseline}C. Due to spurious sources/sinks and non-smoothness, the vector projection baseline is not suitable for downstream robotics tasks. Moreover, we show that even if there are no spurious sources/sinks (since the vector projection baseline ignores the geodesic paths on the surface during the diffusion), the approach fails to capture the global symmetry of the object, leading to undesired behavior for the slicing task (Fig.~\ref{fig:tangent_projection_baseline}, iii).

\subsubsection{Euclidean Diffusion Baseline}
\begin{figure}[]
\centering
\includegraphics[width=0.7\linewidth]{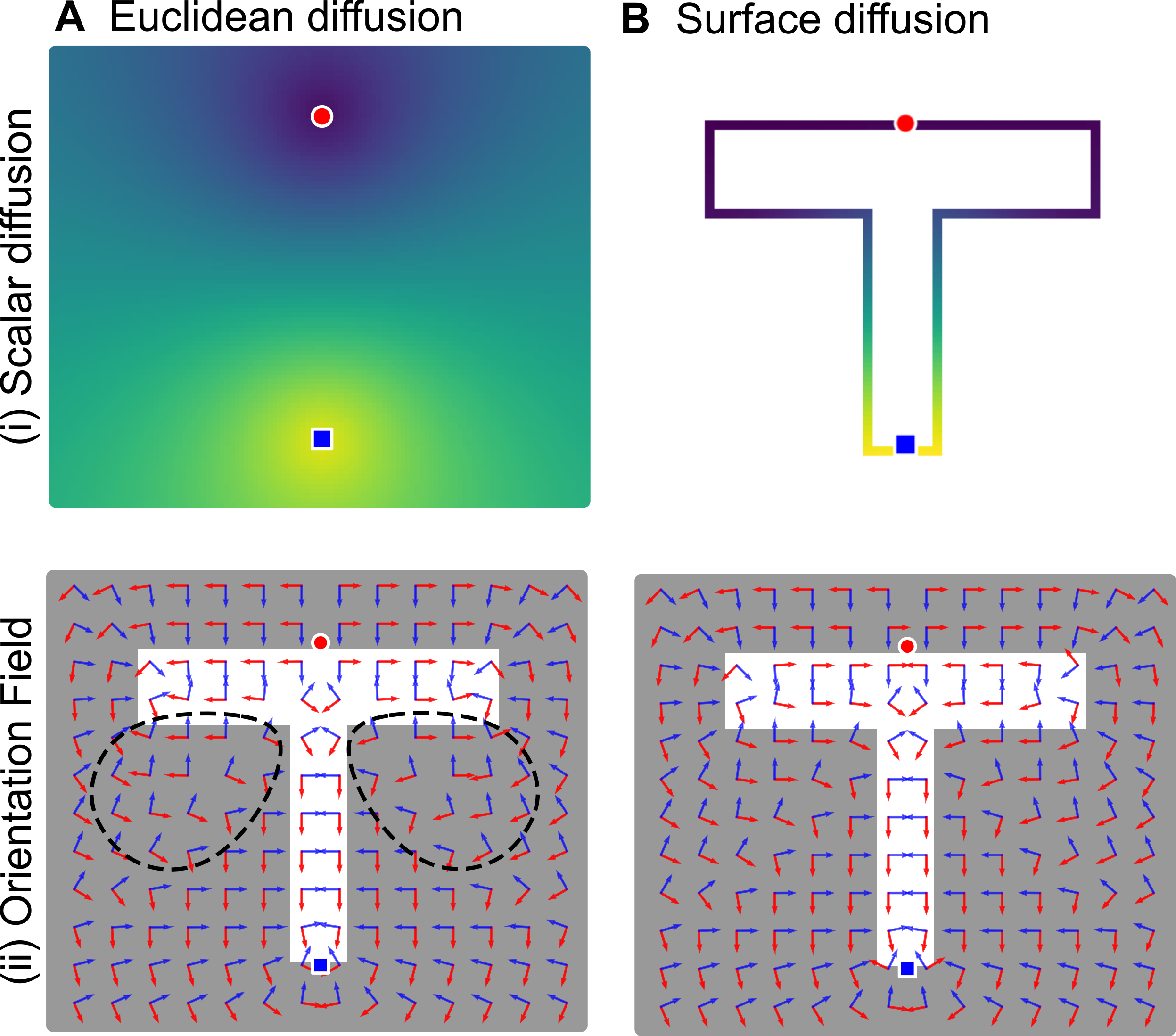}

\caption{\textbf{Comparison between Euclidean diffusion baseline and surface conditioned workspace diffusion}. Red point designates the source whereas the blue square is the sink. (\textbf{A}) Euclidean diffusion baseline, ignoring the surface geometry during the diffusion and using projection of the gradient vectors to the tangent space. (\textbf{B}) Diffused orientation field (ours). (i) Results of scalar Euclidean and surface diffusion. (ii) Orientation field comparison. Regions where the Euclidean diffusion baseline cuts across the concavity of the T-shape are highlighted with dashed lines.}
\label{fig:euclidean_diffusion_baseline}
\end{figure}

As the third baseline, we considered Euclidean diffusion using keypoints (ignoring the surface geometry during the diffusion), computing its gradient field to get x-direction and projecting the resulting vectors to the tangent space. For computing the tangent space we used smooth extension of surface normals to the workspace using WoS. This baseline addresses smoothness issues related to the first two baselines. However, it ignores the geometry of the surface during the diffusion of the x-direction and extrinsically correct it afterwards. We used a T-shaped object for this comparison since it has a concave region which shows the issue created by ignoring the surface geometry for diffusing x-directions. The Euclidean diffusion baseline produces x-directions that cut across the concavity of the T-shape (Fig.~\ref{fig:euclidean_diffusion_baseline}A). This is an undesired behavior since it would lead to undesired motion for downstream robotics tasks. On the other hand, our method respects the geometry of the surface during diffusion and produces x-directions that follow the surface layout (Fig.~\ref{fig:euclidean_diffusion_baseline}B).
\subsection{Robustness to Clutter and Occlusions}
\begin{figure}[]
\centering
\includegraphics[width=0.8\linewidth]{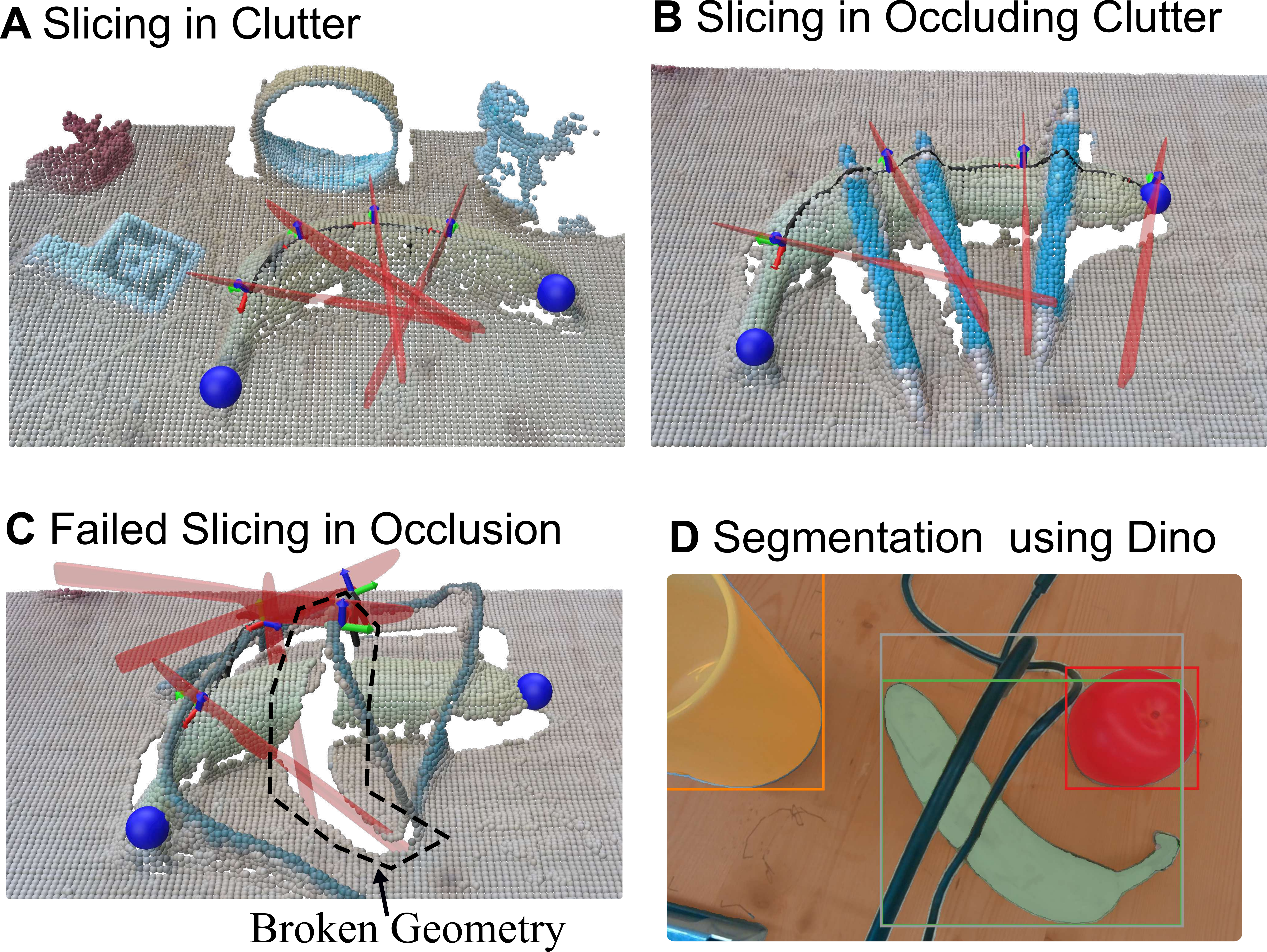}

\caption{\textbf{Expriments on cluttered scenes.} (\textbf{A}) Slicing in the presence of distractor objects. (\textbf{B}) Slicing in the presence of an occluder that does not break the object into disconnected regions. The occluder perturbs local shape but does not break the object's global symmetry, so the trajectory remains stable. (\textbf{C}) Slicing in the presence of an occluder that breaks the object into disconnected regions. We placed a cable across the object so that it occluded the midsection of a banana, producing two disconnected components. With zero-Neumann boundary conditions (which act as diffusion insulators), the diffusion "followed" the cable and led to failure.  (\textbf{D}) Proof-of-concept experiment to address the failure cases in (\textbf{C}) using Grounded SAM~\cite{renGroundedSAMAssembling2024}. We use object segmentation for detecting that the disconnected regions belong to the same object.}
\label{fig:clutter}
\end{figure}

Our method is robust to clutter: objects that are close in Euclidean distance but far in geodesic distance on the target surface have limited influence on the diffusion. Since the diffusion follows geodesics, distractor surfaces do not alter geodesic paths along the object. We show this in a cluttered scene in Fig.~\ref{fig:clutter}A: despite nearby distractors, the slicing trajectory is unaffected.
A second case is occlusion, where the object of interest is partially hidden. If the occlusion does not split the object into disconnected regions, we do not observe problems (Fig.~\ref{fig:clutter}B). Although occluders perturb local shape, they do not break the object's global symmetry, so the trajectory remains stable. Naturally, if the occluder dominates the view, performance degrades. In such cases, the occluder should first be removed or the camera viewpoint changed.
A problematic case arises when an occluder breaks the observed geometry into disconnected regions (Fig.~\ref{fig:clutter}C). We placed a cable across the object so that it occluded the midsection of a banana, producing two disconnected components. With zero-Neumann boundary conditions (which act as diffusion insulators), the diffusion “followed” the cable and led to failure. A potential fix is to add a light instance-segmentation step (e.g., to detect that the disconnected components belong to the same object), then treat internal vs. external boundaries differently—only assigning zero-Neumann to external boundaries. This preserves diffusion across the object while insulating against true external occluders. We made a successful proof-of-concept test for this approach using Grounded SAM~\cite{renGroundedSAMAssembling2024} for segmenting the objects and figuring out the disconnected regions belong to the same object, as shown in Fig.~\ref{fig:clutter}D. 
\subsection{Force and Contact Stability}
Point cloud measurements have millimeter scale position errors from the combined effects of camera depth noise, residual errors in extrinsic calibration, and joint encoder uncertainty. While the position from the point cloud can be biased by a few millimeters, the local directions from DOF remain reliable for approaching or moving on the surface, and the admittance controller compensates positional offsets during contact-rich tasks such as peeling.

\begin{figure}[]
\centering
\includegraphics[width=1.0\linewidth]{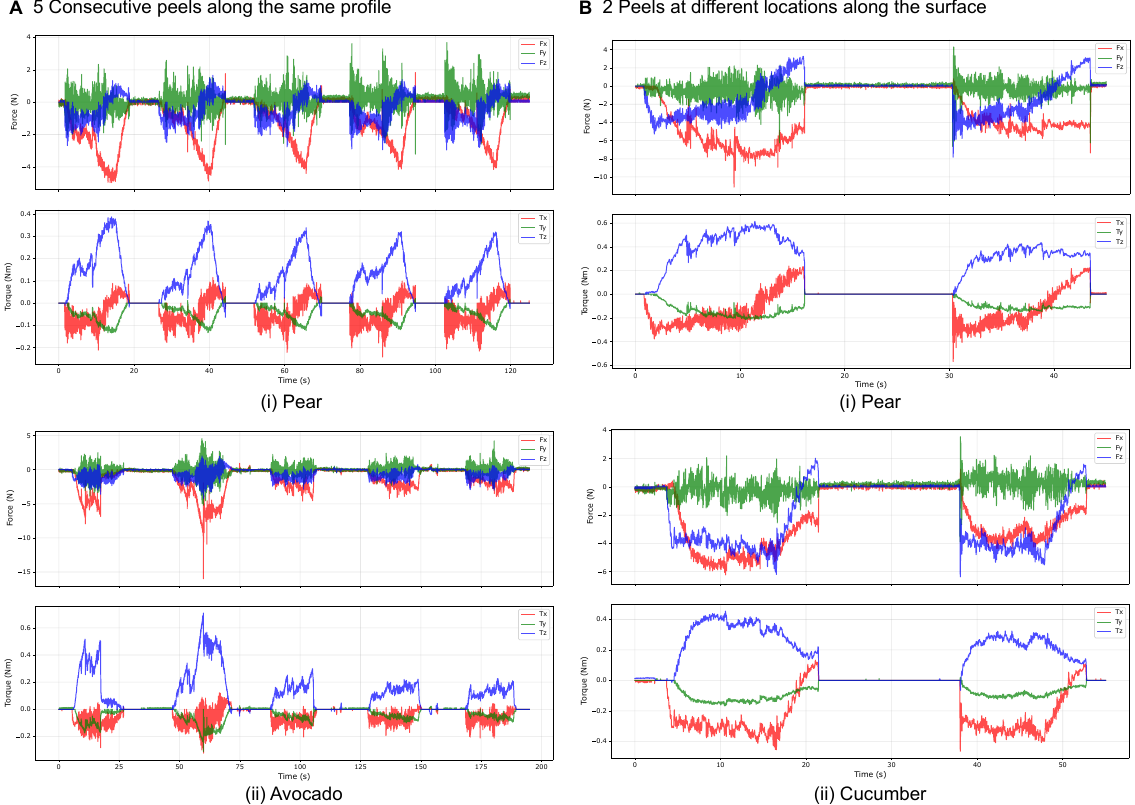}

\caption{\textbf{Force and torque profiles from real-world peeling experiments.} Forces and torques are expressed in the end effector frame which is aligned with the local reference frame. (\textbf{A}) 5 consecutive peels from (i) a pear and (ii) an avocado,  by removing progressively more material and effectively changing the surface height. (\textbf{B}) 2 peels at different locations along the surface of (i) a pear and (ii) a cucumber.}
\label{fig:force_profiles}
\end{figure}

We evaluated the force and contact stability in two experiments. The first shows adaptation to height error by performing five consecutive peels that progressively lower the surface. The second reports force and torque profiles across objects. Figure~\ref{fig:force_profiles}A plots five consecutive peels for pear and avocado. For the pear, force and torque traces are closely matched across passes; notably the z direction force becomes positive near the end of the trajectory due to the cutting force required to separate the skin. We also observe a small decrease in normal force and tool axis torque across passes, consistent with progressive material removal. For the avocado, the skin detaches mid way through the first peel, seen as the x direction force dropping to zero. In the second run, the peeler re-engages at the remaining unpeeled patch and continues peeling from that point. After the hard skin is removed, later peels run on soft flesh and produce lower forces and torques with similar profiles.

Figure~\ref{fig:force_profiles}B shows two peels at different locations on the same object for pear and cucumber. Profiles are similar in shape, with magnitudes that vary across objects. For all the experiments, we used the same hyperparameters listed in Table~\ref{tab:perception_params}. Measurement precision is limited by fixture compliance, since rigidly securing soft objects without deformation is difficult and beyond the scope of this work.

\begin{table}
  \centering
  \caption{\textbf{Hyperparameters.}}
  \label{tab:perception_params}
  \begin{tabular}{lll}
    \hline
    \textbf{Group} & \textbf{Parameter} & \textbf{Value} \\
	    \hline
				& Diffusion Time ($\tau$) 		& 1000 (10 only for coverage) \\
				& Voxel Downsampling    & 3 mm \\
		\hline
    Admittance Controller 				
           		& Position Stiffness (N/m) 			& [400,200,200] \\
           		& Rotational Stiffness (N*m/rad)	& [200,200,200] \\
    \hline
  \end{tabular}
\end{table}


\clearpage 





\end{document}